\documentclass{article} 
\usepackage{iclr2025_conference,times}

\usepackage{amsmath,amsfonts,bm}

\def\eqref#1{equation~\ref{#1}}

\def\1{\bm{1}}

\DeclareMathAlphabet{\mathsfit}{\encodingdefault}{\sfdefault}{m}{sl}
\SetMathAlphabet{\mathsfit}{bold}{\encodingdefault}{\sfdefault}{bx}{n}

\usepackage{hyperref}
\usepackage{url}
\usepackage{pifont} 

\PassOptionsToPackage{table}{xcolor}
\usepackage[most]{tcolorbox}

\usepackage[table]{xcolor}

\usepackage{microtype}
\usepackage{graphicx}
\usepackage{subfigure}
\usepackage{booktabs} 
\usepackage{enumitem}
\usepackage{listings}
\usepackage{array}

\newtcolorbox[auto counter, number within=section, list type=subsubsection, list inside=toc]{sectionbox}[2][]{
colback=white!98!gray, colframe=black, 
colbacktitle=white!90!gray, coltitle=black, 
fonttitle=\bfseries,
title={#2}, 
list entry={Comment \thetcbcounter\quad}
}

\usepackage{amsmath}
\usepackage{array}
\usepackage{booktabs} 

\newcommand{\method}{\textsc{StepWiser}}

\title{StepWiser: Stepwise Generative Judges for Wiser Reasoning}

\author{Wei Xiong$^{1,2}$, Wenting Zhao$^1$, Weizhe Yuan$^{1,3}$, Olga Golovneva$^1$, Tong Zhang$^2$, \\{\bf Jason Weston$^{1,3}$, Sainbayar Sukhbaatar$^1$}\\
$^1$FAIR at Meta, $^2$University of Illinois Urbana-Champaign, $^3$NYU}

\iclrfinalcopy 
\begin{document}

\maketitle

\begin{abstract}
As models increasingly leverage multi-step reasoning strategies to solve complex problems, supervising the logical validity of these intermediate steps has become a critical research challenge. Process reward models address this by providing step-by-step feedback, but current approaches have two major drawbacks: they typically function as classifiers without providing explanations, and their reliance on supervised fine-tuning with static datasets limits generalization.
Inspired by recent advances, we reframe stepwise reward modeling from a classification task to a reasoning task itself. 
We thus propose a generative judge that {\em reasons about the policy model's reasoning steps} (i.e., meta-reasons), outputting thinking tokens before delivering a final verdict. 
Our model, \method{}, 
is trained by reinforcement learning using relative outcomes of rollouts.
We show  it provides (i) better judgment accuracy on intermediate steps than existing methods; (ii) can be used to improve the policy model at training time; and (iii) improves inference-time search.
\end{abstract}

\section{Introduction}

As large language models (LLMs) increasingly tackle complex problems, they rely on multi-step reasoning strategies like Chain-of-Thought (CoT) \citep{wei2022chain} and ReAct \citep{yao2022react} to decompose tasks and formulate better solutions. Consequently, ensuring these intermediate reasoning steps possess logical validity has become a critical research challenge. Process Reward Models (PRMs) have emerged as a potential tool to meet this need, providing step-by-step feedback for supervising learning, instead of relying on a single, often sparse, outcome-based reward \citep{lightman2023let, wang2023math}. However, this approach suffers from two major drawbacks. First, current PRMs typically function as ``black-box'' classifiers, providing a score or label without explaining why a step is correct or flawed. Second, their reliance on supervised fine-tuning (SFT) with static datasets can limit their ability to generalize to new reasoning patterns \citep{lightman2023let, luo2024improve, wang2023math, xiong2024implementation, zhang2024entropyregularizedprocessrewardmodel}.
In contrast, reasoning models themselves are trained to produce CoTs with reinforcement learning (RL) for best performance \citep{deepseekai2025deepseekr1incentivizingreasoningcapability}.

\begin{figure}
    \centering
    \includegraphics[width=1\linewidth]{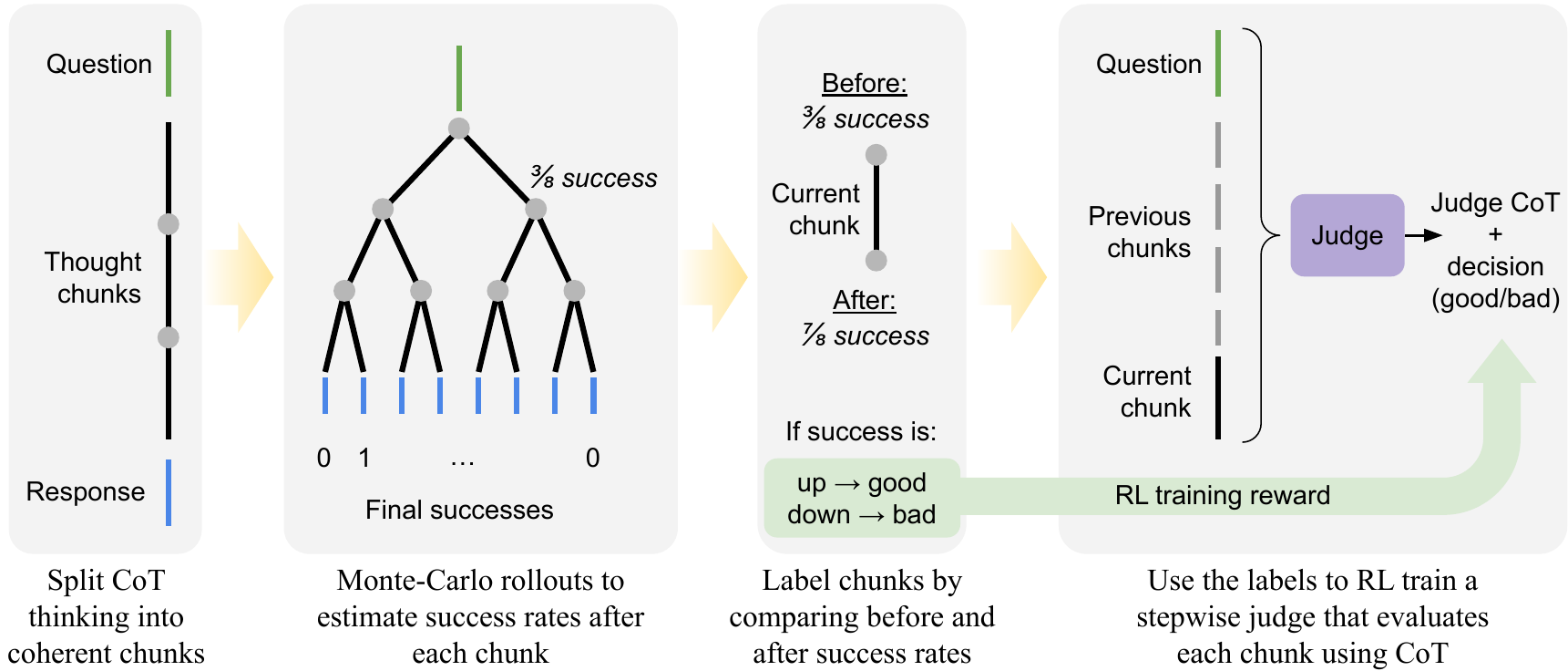}
    \caption{\textbf{Overview of our \method{} training method:} we teach the model to segment its chain-of-thought (CoT) into coherent chunks. Then after each chunk, we generate Monte-Carlo rollouts to estimate the average success rate (i.e.~Q-value) starting from that point. If the success rate goes up (or down) after a given chunk, we label it as good (or bad). Using these labels, we RL train a stepwise judge model that determines the quality of a given chunk after its own CoT reasoning.}
    \label{fig:method}
\end{figure}

In this paper we propose to reward intermediate reasoning steps by first {\em\bf reasoning about those reasoning steps}, before making a judgment -- a meta-reasoning process which itself is trained by RL.
Our overall method (as shown in \autoref{fig:method}) to build such a {\em stepwise generative  judge}  involves 3 components: (1) a new self-segmentation technique to equip the base policy model with the ability to produce coherent and informative reasoning chunks (chunks-of-thought); (2) assignment of target rewards to chunks via relative outcomes of rollouts; and
(3) online training of judgment reasoning chains (i.e., reasoning about reasoning) and final reward judgments via RL. Our stepwise judge, termed \method{}, can then be used to provide rewards either at training time or inference time in order to improve the reasoning ability of the policy model.

We conduct a comprehensive evaluation of our method across three key dimensions: (i) the judge's classification accuracy on intermediate steps, e.g., via its score on ProcessBench \citep{zheng2024processbench}; (ii) its performance in a new inference-time search paradigm where the judge cleans up the reasoning history and re-samples -- a method we propose for efficiently scaling sequential computation while maintaining the original generation length;   and (iii) its utility in data selection for downstream model training. Our experiments demonstrate that our RL-trained generative stepwise judge significantly outperforms traditional SFT-based baselines and other existing methods across all axes of evaluation, where  the ability to meta-reason  -- trained via RL -- is the critical factor.

\section{Related Work} 

\subsection{Process Reward models in LLM math reasoning} To improve the reliability of multi-step reasoning in LLMs, one can consider methods beyond 
evaluating only the final answer, termed Outcome Reward Models (ORMs), by instead evaluating each intermediate step, a method pioneered by Process Reward Models (PRMs). \citet{lightman2023let} first demonstrated that a process-supervised model can significantly outperform an outcome-supervised one in guiding best-of-n sampling. However, their PRM800K dataset relied on intensive human annotation for each reasoning step, which is generally infeasible for larger, more diverse and challenging datasets.

Subsequent research has focused on automating this annotation process. \citet{wang2023math} proposed using Monte Carlo (MC) rollouts to estimate the Q-value of each step, while \citet{luo2024improve} introduces a binary search method to efficiently identify faulty steps. Our work builds upon the MC-based annotation approach, exploring various methods for converting these Q-value estimates into effective learning signals.

In parallel, another line of work has established a theoretical connection between intermediate step values and the final outcome within the framework of KL-regularized Markov Decision Processes \citep{zhong2024dpo, rafailov2024r}. This result has been used to derive DPO-like objectives for learning an implicit PRM from outcome-only data \citep{xiong2024building, cui2025processreinforcementimplicitrewards,zhou2025sweet} or a KL-regularized version of the MC-based estimator \citep{zhang2024entropyregularizedprocessrewardmodel}. A recent work \citep{zha2025rl} prompts LLMs to evaluate each individual step before producing a final judgment, but supervises only the evaluation of the final answer. These methods share the goal of learning a stepwise judge from sparse signals. The major advantage of this line of work is that it is easy to train the judge in an online manner. A central question, which our work addresses, is whether the rich, explicit signals from extensive Monte Carlo sampling provide a more effective learning signals for RL-based reward training. 

Concurrent work by \citet{he2025good} uses a prompting approach to segment thought process into coherent chunks similar to ours. However, their stepwise judge is based only on prompting techniques that leverages hints in CoT like ``Wait, I made a mistake''.
In contrast, our method focuses on training a judge using  stepwise labels grounded in final verified answers.

\subsection{Judge architectures} 

The process rewards described above can be used to train judges with different distinct architectures and training paradigms.

\paragraph{Discriminative PRMs} The most straightforward approach is to treat the task as a {\em classification problem}. This involves replacing the language model's final layer with a linear head and fine-tuning it to predict a binary label for each step using a cross-entropy loss \citep{lightman2023let}. A more recent method formulates the task as next-token prediction, prompting the LLM to generate a pre-defined token (e.g., + or -) as its judgment \citep{wang2023math, xiong2024implementation}. This approach further dates back to preference reward model training \citep{dong2024rlhf, liu2023statistical}. Although this method uses a generative mechanism, its function remains purely discriminative, as it outputs a simple judgment without justification. We therefore group both under the discriminative category.

\paragraph{Generative judges with CoT reasoning} In sharp contrast, the second and most recent paradigm is the generative reasoning judge. Here, the evaluation itself is framed as a reasoning task. The judge first generates an explicit CoT to explain its rationale before outputting its final judgment. This approach was initially explored for preference learning and ORMs \citep{zhang2024generative, chen2025rm}. There are also a few very recent works studying this paradigm shift in the context of stepwise judges, including \citet{zhao2025genprm, zha2025rl, khalifa2025process}. Though we share similar spirit of leveraging the inherent reasoning ability of the LLMs to train a stepwise judge, the algorithmic designs are distinctly different. 

In contrast to \citet{zhao2025genprm} and \citet{khalifa2025process}, who focus on offline rejection sampling fine-tuning, our investigation shows that such static training methods suffer from scalability issues, where the performance quickly stagnates after the initial few steps. In contrast, we cast the stepwise judgment as a reasoning task, and focus on online RL training.  \citet{zha2025rl} do use RL, but with sparse, trajectory-level supervision. Specifically, they prompt the LLMs to evaluate each individual step and final answer but only the final verification is supervised. Their approach assumes that to get an accurate evaluation of the final answer, models \textit{implicitly become a stepwise judge}. In contrast, our framework is built on dense, stepwise supervision via rollouts. Our experiments will show that this explicit, online, stepwise signal is critical for training state-of-the-art generative judges.

\section{Method: Training stepwise generative judges with RL}

As depicted in \autoref{fig:method}, our overall method \method{} consists of three components: 
\begin{itemize}
\item We equip the base policy model with the ability to self-segment Chain-of-Thoughts into coherent and informative reasoning chunks, called Chunks-of-Thought. This is done by creating SFT data with informative segments, so that the model can be trained to self-segment. We show that this causes no loss in performance for the base model.
\item Given the chunks generated by the policy model, we annotate each chunk to create training data for our generative stepwise judge with binary target labels. This is done by comparing outcomes of rollouts starting before and after the given chunk using the outcome rewards.
\item We perform online RL training using GRPO which trains our stepwise judge model to produce  judgment reasoning chains (i.e., reasoning about reasoning) and reward final judgments that match the chunk labels from the previous step.
\end{itemize}

We describe  the three components  in detail in the following three subsections.

\subsection{CoT Generation with Self-Segmentation (Chunks-of-Thought)}

To train judges that can evaluate individual steps in a reasoning process, a key challenge is defining what a ``step'' is. While CoT reasoning enables models to reason step by step, properly segmenting this reasoning remains a difficult problem.

Current methods often segment reasoning trajectories using pre-defined tokens, like ``Step 1, Step 2'' or simply using double line breaks as delimiters. However, these heuristics frequently result in segments that are neither logically complete nor self-contained. Each segment contains only limited information, making it unsuitable as a standalone unit for a judge model to evaluate effectively. We present a representative example in \autoref{tab:chunking} (left), where the model tends to insert double line breaks before and after a mathematical equation.  This breaks an intuitively unified logical step into multiple different chunks, where one chunk contains a textual explanation, and the next with the corresponding equation.

\begin{table*}[t!]
  \caption{Rules that we provide for an LLM to create segmented Chunks-of-Thought SFT data.
   \label{tab:prompt_chunking_list}}
  \small
\centering
\begin{tcolorbox}[
    colback=gray!5,
    colframe=gray!60!black,
    title=Rules for CoT Trajectory Segmentation,
    fonttitle=\bfseries,
    sharp corners,
    boxrule=1pt,
]
\textbf{Segmentation Principles}
\begin{enumerate}[label=\arabic*., topsep=3pt, itemsep=2pt, leftmargin=*]
    \item Unified purpose: A chunk should serve a single, clear objective. For example: setting up an initial equation, executing a self-contained calculation (like integration by parts), or stating a final/intermediate conclusion. All content within the chunk must directly serve this one core goal.
    \item Logical Cohesion: All lines within a chunk must form a continuous and uninterrupted logical flow. A new chunk should begin as soon as the focus or purpose of the reasoning shifts.
    \item Clear Transition: A new chunk must begin when the problem-solving process enters a new phase. This includes transitioning from "solving for a variable" to "verifying the answer," or inserting an "explanatory side-note" into the main workflow.
    \end{enumerate}
    \textbf{Format rules.}
\begin{enumerate}[label=\arabic*., topsep=3pt, itemsep=2pt, leftmargin=*]
    \item Use \texttt{\textless chunk\textgreater \dots\ \textless/chunk\textgreater} to mark the beginning and end of each segment. The text and newlines inside the tags must not be altered.
    \item The final output should only contain the tagged content, without any additional text, titles, or blank lines.
    \item You \textbf{must} preserve all original text and newlines exactly as they appear within the tags.
\end{enumerate}
\end{tcolorbox}
\end{table*}

\paragraph{Achieving better step definition via self-segmentation} To mitigate this issue, we propose a method to teach the model to generate and simultaneously self-segment its own reasoning chains into more meaningful steps. First, we define the criteria for a high-quality reasoning step. The core idea is that each step should represent a complete logical leap or a self-contained part of the problem-solving process. Our definitions are given in~\autoref{tab:prompt_chunking_list}. We then create our training data by:
\begin{enumerate}
    \item Generating a set of initial reasoning trajectories from the base model.
    \item Using an LLM prompted with our rules, to automatically segment these trajectories into logically coherent steps.
\end{enumerate}


We fine-tune our base model on this data, thus teaching it to generate and simultaneously self-segment its own reasoning chains automatically. This self-segmentation ability is crucial for two main reasons. First, it produces more informative and logically complete steps, which provides better context for our judge model and improves its evaluation accuracy. Second, this method significantly reduces the total number of steps per trajectory. This reduction is also important because, as we will show, the process of annotating each step with a quality label is computationally expensive.

\begin{table*}
    \caption{A representative example illustrating CoT segmentation methods based on a response from Qwen2.5-1.5B-it. Standard splitting by newlines (left) results in too many chunks, often separating textual explanations from their corresponding mathematical equations. In contrast, our self-segmentation approach (right) yields more coherent and meaningful chunks, with a clear, self-contained purpose—simplifying the second term in the previously established equation. Additional full examples are provided in the appendix. \label{tab:chunking} }
\vspace{6pt}
\centering
\tiny
\begin{tabular}{|m{0.47\textwidth}|m{0.47\textwidth}|}
\hline
\centering\textbf{Split by \texttt{\textbackslash n\textbackslash n}} & \centering\textbf{LLM-based self-segmentation (Chunks-of-Thought)} \tabularnewline
\hline
...

\vspace{6pt}
\textbf{chunk 9} \par
Similarly, for the second term, we have:

\vspace{6pt}
\textbf{chunk 10} \par
\[
\left|\frac{z^2 + 4i - 3}{z - i + 2}\right| = \left| \frac{(x+yi)^2 + 4i - 3}{x+yi - i + 2} \right|.
\]
\vspace{6pt}
\textbf{chunk 11} \par
Since \( z^2 + 4i - 3 = x^2 - y^2 + 2xyi + 4i - 3 = (x^2 - y^2 - 3) + (2xy + 4)i \), we get: 

\vspace{6pt} 
\textbf{chunk 12} 
\[
z - i + 2 = x + (y-1) = x + yi - i + 2 = (x + 2) + yi.
\]
\vspace{6pt} \textbf{chunk 13} \par
Using the same logic as above, the magnitude is:
\vspace{6pt} \par
...

 & 

...

\textbf{chunk 3} \par
Similarly, for the second term, we have:
\[
\left|\frac{z^2 + 4i - 3}{z - i + 2}\right| = \left| \frac{(x+yi)^2 + 4i - 3}{x+yi - i + 2} \right|.
\]
Since \( z^2 + 4i - 3 = x^2 - y^2 + 2xyi + 4i - 3 = (x^2 - y^2 - 3) + (2xy + 4)i \), we get:
\[
z - i + 2 = x + (y-1) = x + yi - i + 2 = (x + 2) + yi.
\]
Using the same logic as above, the magnitude is:
$$
\begin{aligned}
    &\sqrt{\left(x^2 - y^2 - 3 + x + y + 2\right)^2 + \left(2xy + 4 + x + y\right)^2} \\&= \sqrt{\left(x^2 - y^2 - 1 + x + y\right)^2 + \left(2xy + x + y + 4\right)^2}.
\end{aligned}
$$

\textbf{chunk 4}

...

\tabularnewline
\hline
\end{tabular}
\end{table*}

\subsection{Stepwise Data Annotation} \label{sec:mcq}

\paragraph{Stepwise data annotation via Q value estimation} Previous work has used human labelers to annotate correctness of each reasoning step \citep{lightman2023let}, although most such data is collected  for proprietary models that we cannot access. Other works annotate steps automatically using methods like Monte Carlo estimation \citep{wang2023math}. We follow this second approach, using an estimated Q-value to measure the quality of each step.

For a given training prompt $x$ with verifiable outcome rewards, we generate a response from our policy model $\pi$ which segments its CoT into chunks $a = [a_1, a_2, \cdots, a_H]$, where $a_i$ is the i-th reasoning chunk. Then, the Q value of an individual step $a_i$ and its history is the expected final reward starting from that point:
\begin{equation}
    Q^\pi\big([x, a_{1:i-1}], a_i\big) :=  Q^\pi(s_{i-1}, a_i) = \mathbb{E}_{a_{i+1 : H} \sim \pi(\cdot|x, a_{1:i})} r^\star(x, a_{1:H}),
\end{equation}
where $s_i := [x, a_{1:i-1}]$ is the history, and $r^\star$ is a final reward, which can be 1 for correct answers and 0 otherwise. We estimate this Q-value by generating $M$ full completions $a_{i+1:H}^j$ from that step $a_i$ and calculating the average final reward, i.e.\ the ratio of correct final answers:
\begin{equation}
\label{eq:q_est}
    \widehat{Q}^\pi \big(s_{i-1}, a_i\big) = \frac{1}{M} \sum_{j = 1}^M r^\star(x, a_{1:i}, a_{i+1:H}^j).
\end{equation}
Following prior work \citep{wang2023math, xiong2024implementation}, we can then assign a binary label to the step based on this Q-value:
$$
\begin{aligned}
    y_i = 
\begin{cases}
+ & \quad \text{if } \widehat{Q}^\pi \big(s_{i-1}, a_i\big) > 0, \\
- & \quad \text{if } \widehat{Q}^\pi \big(s_{i-1}, a_i\big) = 0.
\end{cases}
\end{aligned}
$$
For convenience, we refer to this labeling approach as  \textit{Absolute Q value thresholding} (\texttt{Abs-Q}).

\paragraph{Rewarding the progress} One drawback of \texttt{Abs-Q} is its insensitivity to the dynamics of the reasoning process. For instance, it does not differentiate between a step that raises the success probability from 10\% to 50\% and one that drops it from 60\% to 55\%. To reward progress, we also explore methods that consider the change in Q-value.

\citet{setlur2024rewarding} proposes to consider the change in value. Specifically, they define the notion of \textit{effective reward} as a combination of Q value and advantage function of the best-of-n policy induced by $r^\star$:
\begin{equation}
    Q^\pi(s_{i-1}, a_i) + \alpha \cdot A^\mu (s_{i-1}, a_i), 
\end{equation}
where $\alpha > 0$ is a hyperparameter, and $A^\mu(s_{i-1}, a_i) := Q^\mu(s_{i-1}, a_i) - Q^\mu(s_{i-2}, a_{i-1})$. Here $\mu$ is taken as the best-of-n policy with $r^\star$. In other words, we generate $n$ responses from $\pi$ and use $r^\star$ to select the best one. In this case, $\mu$ satisfies that $Q^\mu(s_{i-1}, a_i) = 1 - (1- Q^\pi(s_{i-1}, a_i) 
)^n$\footnote{Assuming binary $\{0,1\}$ outcome rewards where $(1- Q^\pi(s_{i-1}, a_i) 
)^n$ is the probability of $n$ rollouts failing.}. Therefore, the effective reward can also be estimated via Q value estimation. Accordingly, we consider an alternative approach of data annotation:
$$
\begin{aligned}
    y_i = 
\begin{cases}
+ & \quad \text{if } \widehat{Q}^\pi\big(s_{i-1}, a_i\big)  + \alpha \cdot \widehat{A}^\mu \big(s_{i-1}, a_i\big) > 0, \\
- & \quad \text{if } \widehat{Q}^\pi\big(s_{i-1}, a_i\big) + \alpha \cdot \widehat{A}^\mu \big(s_{i-1}, a_i\big) = 0,
\end{cases}
\end{aligned}
$$
where $\widehat{A}^\mu$ is the estimated advantage through the Monte-Carlo estimation of the Q value. We refer to this labeling approach as \textit{Relative Effective Reward Thresholding} (\texttt{Rel-Effective}).

As a simpler alternative to capture relative improvement, we also consider a method based on the value ratio, where the label is determined as:  
$$
\begin{aligned}
    y_i = 
\begin{cases}
+ & \quad \text{if } \widehat{Q}^\pi\big(s_{i-1}, a_i\big) / \widehat{Q}^\pi\big(s_{i-2}, a_{i-1}\big)> \gamma, \\
- & \quad \text{if } \widehat{Q}^\pi\big(s_{i-1}, a_i\big) / \widehat{Q}^\pi\big(s_{i-2}, a_{i-1}\big) \leq \gamma.
\end{cases}
\end{aligned}
$$
Here $\gamma > 0$ is a threshold and we refer this labeling approach as \texttt{Rel-Ratio}.

Using one of these methods, we can assign binary label $y_i$ to every step $a_i$ in a reasoning trajectory.
Since these labels come from unbiased estimates of the actual Q-values, they are likely to be more reliable compared to more ad-hoc methods.
For example, if a step $a_i$ is the first step with a mistake, rollouts starting after $a_i$ are more likely to fail compared to ones that start before the flawed step $a_i$. 

\subsection{Training the Judge via RL}

At this stage we now have the recipe to create segmented (chunked) reasoning chains, each with a stepwise target label, across our training data. A straightforward approach would be to train a judge model as a classifier using standard SFT, as done in prior works \citep{wang2023math, xiong2024implementation}. However, recent studies suggest that a more robust and effective (in their case, outcome-based) judge can be created by having it generate its own CoT analysis to evaluate the models' responses \citep{zhang2024generative, chen2025rm, whitehouse2025j1}. In the meantime, this generative formulation naturally allows us to train a judge via reinforcement learning. We therefore frame the stepwise evaluation as a reasoning task where the judge model first generates an analytical rationale and then concludes with a final judgment. This approach is compelling also because it forces the judge to ``show its work'', providing a more transparent and potentially more accurate evaluation process.

\paragraph{Task formulation and prompt dataset balancing} We decompose the full trajectories into step-level training prompts (judgment tasks). For each training prompt, the model is provided with the \textit{original problem} $x$, the \textit{reasoning history} $a_{1:i-1}$, and the \textit{new reasoning chunk} $a_i$ to be evaluated. The model is prompted to generate its own CoT reasoning about the correctness of the step $a_i$, followed by a final judgment in a predefined format (e.g., enclosed in a box). Such CoT reasoning in the judge allows it to spend more compute and perform thorough analysis of a reasoning step $a_i$, which is likely necessary given $a_i$ itself is a part of CoT that performs non-trivial reasoning. See Table~\ref{tab:judge_prompt} for the prompt template used. 

A critical but often overlooked aspect of the stepwise judge training is that the stepwise labels can be highly imbalanced due to the data annotation process in Section~\ref{sec:mcq}. For example, with the Qwen2.5-1.5B-chunk model, $70.2\%$ of \texttt{Abs-Q} samples are labeled as correct. In our early experiments, we observe that these imbalanced training prompts can cause model degeneration, as the model can achieve a high score by simply predicting ``correct'' in most cases.  To mitigate this, we make \emph{prompt dataset balancing} an explicit part of our method: we down-sample the majority class so that the numbers of positive and negative prompts are equal. We find this balancing step to be essential for stable RL training, as it ensures the reward signal reflects the model’s discriminative ability rather than class frequency bias. We will study its impacts on the final model performance in Section~\ref{sec:understand_performance}.

\paragraph{Reward and RL training} The training signal for RL is direct and intuitive. For each step $a_i$, the judge model receives a reward of $1$ if its judgment aligns with the label $y_i$ (created by Monte-Carlo estimations), and $0$ otherwise. We use GRPO \citep{shao2024deepseekmath} as our optimization algorithm due to its demonstrated effectiveness across multiple studies \citep{shao2024deepseekmath, deepseekai2025deepseekr1incentivizingreasoningcapability}.

\begin{table}[t!]
\small
  \centering
    \caption{Prompt Template for our \method{} judge.}
\begin{tcolorbox}[
    colback=gray!5,
    colframe=gray!60!black,
    title=Prompt Template for \method{} Judge,
    fonttitle=\bfseries,
    sharp corners,
    boxrule=1pt,
]
\textbf{Instruction:} \\
You are a reasoning validator for mathematical problems. Your task is to think step by step and determine if the ``New Reasoning Chunk'' contains \textbf{any explicit errors} based on the problem description and historical context.

First, you must \textbf{always} perform a step-by-step chain of thought analysis to justify your final judgment. Then, based on your analysis, you will make a definitive judgment. It is OK that the chunk does not contain any numerical calculation.

Based on your evaluation, provide your final judgment:
\begin{itemize}[label=\textbullet, leftmargin=*, topsep=2pt, itemsep=1pt]
    \item Use \textbf{Positive} if the reasoning chunk is free of mistakes.
    \item Use \textbf{Negative} if the reasoning chunk contains one or more mistakes.
\end{itemize}

\vspace{0.5em}\hrule\vspace{1em}

Input:\\
Mathematical Problem:~~\textcolor{orange}{\texttt{\{problem\}}}\\
Historical Reasoning Path:~~\textcolor{orange}{\texttt{\{history\}}}\\
New Reasoning Chunk:~~\textcolor{orange}{\texttt{\{chunk\}}}\\

\vspace{0.25em}\hrule\vspace{1em}

\textbf{Output format:}
\begin{enumerate}[label=\arabic*., topsep=3pt, itemsep=2pt, leftmargin=*]
    \item \textbf{Analysis}: [Always provide a step-by-step analysis here. First, briefly state the goal of the current reasoning chunk. Second, verify the logic, method, and any calculations against the problem's requirements and the historical path. If an error is found, clearly explain the error and why it's wrong. If the reasoning is correct, explain why it is a valid and logical step forward.]
    \item \textbf{Final Judgment}: [Provide the final judgment within \texttt{\textbackslash boxed\{\}}. Examples: \texttt{\textbackslash boxed\{Positive\}} or \texttt{\textbackslash boxed\{Negative\}}.]
\end{enumerate}
\label{tab:judge_prompt}\vspace{2pt}
\end{tcolorbox}
\end{table}

\section{Experiments}\label{sec:exp_results}

\subsection{Experiment Setup}


\paragraph{Model and data} We conduct experiments with the Qwen2.5-1.5B-it and Qwen2.5-7B-it instruction-tuned models \citep{qwen2} which have context window length of 8192. The ground-truth scores for solution verification are provided by Math-Verify\footnote{\url{https://github.com/huggingface/Math-Verify}}. For our training data, we use mathematical problems from the NuminaMath-CoT dataset \citep{numina_math_7b}. Before training, we preprocess the dataset by first removing duplicate prompts. We then use Math-Verify to extract the final answer from each reference solution and score it against the labeled ground truth. We discard any prompts where Math-Verify cannot successfully verify the answer. Unless otherwise specified, we focus on using the same base model to initialize both the policy and the judge. We  also conduct experiments to study how the choice of base model affects the judge's evaluation ability.

\paragraph{Self-segmentation fine-tuning} To create demonstration data for self-segmentation, we use a random subset of $20$k prompts from NuminaMath-CoT. First, we generate 16 responses for each prompt using the base policy model (i.e., Qwen2.5-1.5B-it). Next, we filter out incorrect responses, keeping up to $4$ correct solutions per prompt.  We then prompt the strong Llama-3.1-70B-it \citep{meta_llama3} to segment these correct responses according to the rules in Table~\ref{tab:prompt_chunking_list}. We generate 8 segmentations for each response and only keep those that perfectly reconstructed the original response and follow the required format.  Finally, we fine-tune the base model on this collected data to create Qwen2.5-1.5B-chunk.  For this fine-tuning, we use the open-source Axolotl package\footnote{\url{https://github.com/axolotl-ai-cloud}} with a learning rate of $1e-5$, a packing block size of $8192$ tokens, and a global batch size of $32$.  The setup for Qwen2.5-7B-chunk is similar. We provide the prompt template we use for the response generation in Table~\ref{tab:prompt_chunking_sft}. Table~\ref{tab:self_segmentation} reports both the average number of steps (evaluated on 10K randomly selected prompts from NuminaMath) and the average@32 performance on MATH500. After self-segmentation fine-tuning, the resulting models achieve comparable or slightly better test accuracy on MATH500, while maintaining similar response lengths. Notably, the number of steps decreases significantly compared to commonly used splitting by \texttt{\textbackslash n\textbackslash n}, from 9.6 to 6.0 with 1.5B model and from 9.9 to 6.8 with the 7B model, suggesting that the models generate responses in a more organized and structured manner. An illustrative example is shown in Table~\ref{tab:chunking}, and additional case studies can be found in the Appendix. Meanwhile, we observe that for most current open-source thinking models that do long reasoning before answering, the number of steps exceeds 150 when trajectories are segmented using \texttt{\textbackslash n\textbackslash n}, with each step containing only about 30 tokens. Due to resource constraints, we do not experiment with these thinking models. However, we expect that our self-segmentation technique would be particularly beneficial in this setting, which we leave for future exploration.

\paragraph{Stepwise data annotation} We select a subset of $40$k prompts from NuminaMath for stepwise data annotation based on a pre-filtering process using the pass@k metric. Specifically, for each prompt, we generate 16 responses using our chunk-tuned models (e.g., Qwen2.5-1.5B-chunk). To ensure the selected prompts are of a suitable difficulty, we filter out prompts where the responses were either all correct or all incorrect. During generation, we use a temperature of 1.0 and set the maximum token limit to $8192$, or until the model produced a final answer. Then, for each intermediate step in a solution, we sample another $M=16$ completions staring from that step for estimating Q-values, as specified in \autoref{eq:q_est}. The intermediate labels are then assigned according to the methods described in Section~\ref{sec:mcq}. Here we mainly follow the annotation framework from previous literature \citep{wang2023math, xiong2024implementation}, which is well-suited to our specific research questions. While advanced engineering—such as employing model ensembles for generation or using powerful LLMs/human for label verification \citep{zhang2025lessons}—may enhance results, these engineering strategies are orthogonal to our main contribution, and we believe they could be integrated for future improvements. The data annotation takes approximately $14$ days on 8 A100 GPUs using the Qwen2.5-7B-chunk model, making it computationally expensive. We notice that self-segmentation fine-tuning significantly reduces the number of chunks during trajectory splitting, thereby saving substantial compute and annotation time. We also present additional ablation results with and without chunking in Appendix~\ref{appendix:chunking}.

\begin{table*}[t!]
    \centering \footnotesize
        \caption{Comparison of the base policy with and without self-segmentation fine-tuning. Overall performance is comparable, but self-segmentation results in less chunks than using split by \textbackslash n \textbackslash n. Here Avg@32 is the test accuracy averaged over 32 trajectories with random seeds.} \vspace{10pt} \label{tab:self_segmentation}
\resizebox{0.85\textwidth}{!}{
    \begin{tabular}{ll|ccc}
    \toprule
\textbf{Generator} & \textbf{Method} & \textbf{\# Steps} & \textbf{\# Tokens} & \textbf{Avg@32 on MATH500} \\
    \midrule 
Qwen2.5-1.5B-it & Split by \textbackslash n\textbackslash n & 9.6 & 686.7 & 44.2 \\
Qwen2.5-1.5B-chunk & Self-segmentation & 6.0 & 714.1 & 44.7\\ \midrule
Qwen2.5-7B-it & Split by \textbackslash n\textbackslash n & 9.9 & 733.0 & 73.3 \\
Qwen2.5-7B-chunk & Self-segmentation & 6.8 & 768.1 & 73.3 \\
    \bottomrule
    \end{tabular}
    }
    \end{table*}

\paragraph{Judge RL training details} We implement the GRPO algorithm using the verl library \citep{sheng2024hybridflow}. During each training step, we use a per-prompt batch size of 1024 and a gradient update mini-batch size of 256. For the GRPO training process, the judge model generates $4$ responses for each prompt. The maximum prompt length is set to $3096$ tokens, and the model can generate up to $3096$ new tokens. The learning rate is set to $1e-6$. 

In our initial experiments, we observe that the model's entropy decreases rapidly, particularly for the 7B model. Since our verification task is a binary classification (correct/incorrect), low entropy causes the model to generate 4 responses with identical final judgments. This leads to zero gradients during the GRPO update, causing performance to get stuck after approximately 200 training steps. To mitigate this issue, we adopt the clip higher technique \citep{yu2025dapo}, using $\epsilon_h = 0.28$ and $\epsilon_l = 0.2$. We expect that more advanced methods, such as those proposed by \citet{lanchantin2025diverse}, could further alleviate this issue.

We apply a heuristic filtering process to remove prompts that were overly short or excessively long. Additionally, we find that our threshold-based labeling method often results in an imbalance between positive and negative examples. Therefore, we downsample the majority class to create a balanced training set to stabilize training. We also include an ablation study on this process. 

We run the training for 800 steps, which takes approximately $5$ days on 8 A100 GPUs (for Qwen2.5-7B-chunk models). We present a typical reward curve and an entropy loss curve in Appendix Figure~\ref{fig:entropy_and_reward}.

\subsection{Evaluation on ProcessBench}

\begin{table*}[t!]
    \centering \scriptsize
        \caption{{\bf ProcessBench results.}   The score of each subset is computed via \autoref{eq:harmonic_mean}. Average accuracy (Avg) of our method \method{} is better than all variants of our discriminative baselines, and existing baselines in the literature (first rows). Further comparisons are given in Appendix Table~\ref{tab:full_process_bench}.} \vspace{10pt} \label{tab:process_bench}
       \resizebox{\textwidth}{!}{%
    \setlength{\tabcolsep}{4pt}
    \begin{tabular}{ll|cccc|c}
    \toprule
 \bf Method & \textbf{Learning signal} & \bf GSM8K & \bf MATH & \bf Olympiad & \bf Omni-MATH & \bf Avg $\uparrow$ \\
 \midrule
 \hspace{-1mm}\textit{Existing Reference Models} & & & & & & \\ 
 Math-Shepherd-PRM-7B&  \texttt{Abs-Q} & 47.9& 29.5 &24.8 &23.8& 31.5\\
 RLHFlow-Llama3-8B-it &  \texttt{Abs-Q} & 50.4& 33.4& 13.8 &15.8 &28.4\\
 Skywork-Qwen2.5-Math-7B-it &  \texttt{Abs-Q} & 70.8 & 53.6& 22.9 &21.0& 42.1\\
Eurus-Qwen2.5-Math-7B-it (DPO) & Outcome  & 56.6 & 43.0 & 27.3 & 26.8 & 35.1 \\
 RL-TANGO-Qwen2.5-7B-it & Outcome  & 53.1 & 48.2  & 37.8 & 36.3 & 43.9 \\
    \midrule 
\hspace{-1mm}\textit{Qwen2.5-1.5B-chunk} & & & & & &  \\ 
Discriminative + SFT & \texttt{Abs-Q} & 39.3 & 32.1 & 19.3 & 18.9 & 27.2 \\
 Discriminative + SFT &  \texttt{Rel-Effective} & 40.8 & 37.2& 18.7 & 20.1 & 29.2 \\
 Discriminative + SFT &  \texttt{Rel-Ratio} & 32.1 & 32.0 & 14.2 & 18.0 & 24.1 \\
 Generative CoT + RL (\method{})&  \texttt{Abs-Q} & 49.2 & 40.5 & 23.8 & 31.0 & 36.1  \\
 Generative CoT + RL (\method{}) &  \texttt{Rel-Effective} & 48.2 & 43.6 &22.1 & 25.3 & 34.8 \\
Generative CoT + RL (\method{})&  \texttt{Rel-Ratio} &  46.9 & 43.4 & 26.3 & 28.4 & 36.2 \\ \midrule
\hspace{-1mm}\textit{Qwen2.5-7B-chunk} & & & & & &  \\ 
Discriminative + SFT & \texttt{Abs-Q} & 54.8 & 45.9 & 28.0 & 26.9 & 38.9 \\
 Discriminative + SFT &  \texttt{Rel-Effective} & 55.6 & 48.7 & 26.4 & 28.3 & 39.7 \\
 Discriminative + SFT &  \texttt{Rel-Ratio} & 48.6 & 46.9 & 21.9 & 25.4 & 35.7\\
 Generative CoT + RL (\method{})& \texttt{Abs-Q} & 61.9 & 61.0 & 48.4 & 43.9 & 53.8 \\ 
      \qquad + Maj@8 & \texttt{Abs-Q} & 65.5 & 62.1  & 49.7 & 45.7 & 55.8 (+2.0)\\
  Generative CoT + RL (\method{})& \texttt{Rel-Effective} & 72.4	& 68.3 &	54.4 & 52.4 & 61.9\\
     \qquad + Maj@8 & \texttt{Rel-Effective} & \bf 72.9 & \bf 72.1 & \bf 57.3 & \bf 54.0 & \bf 64.1 (+2.2)\\
   Generative CoT + RL (\method{}) & \texttt{Rel-Ratio} & 72.6 & 67.2 & 52.3 & 49.8 & 60.5\\
   \qquad + Maj@8 & \texttt{Rel-Ratio} & 74.3 & 69.0 & 53.8 & 50.2 & 61.8 (+1.3)\\ 
    \bottomrule
    \end{tabular}
    }
    \end{table*}

We first evaluate our method \method{} on ProcessBench \citep{zheng2024processbench}, a benchmark designed to test the ability to identify the first incorrect step in a reasoning trajectory, as labeled by human annotators. The benchmark contains 3500 problem-solution pairs from diverse math datasets (GSM8K \citep{cobbe2021gsm8k}, MATH \citep{hendrycks2021measuring}, Olympiad Bench \citep{he2024olympiadbench}, and Omni-MATH \citep{gao2024omni}). Performance is measured by the harmonic mean\footnote{Note that while this score was referred to as the F1 score in the original paper, it is different from the standard F1 score.} of the accuracy on problems with correct final answers $\mathrm{acc}_1$ and those with incorrect final answers  $\mathrm{acc}_2$, calculated as:
\begin{equation}\label{eq:harmonic_mean}
2 \times \frac{\mathrm{acc}_1 \times \mathrm{acc}_2}{\mathrm{acc}_1 + \mathrm{acc}_2}.
\end{equation}


\paragraph{Our RL-trained \method{} judge significantly outperforms SFT-trained discriminative judges} Our primary results on ProcessBench are presented in \autoref{tab:process_bench}. The findings show that our RL-trained \method{} judge significantly outperforms all variants of the SFT-trained discriminative judge. This holds true across all learning signals (\texttt{Abs-Q}, \texttt{Rel-Ratio}, \texttt{Rel-Effective}) and model scales. For instance, on the 7B model with the \texttt{Rel-Effective} signal, \method{} achieves an average score of 61.9, far surpassing the discriminative baseline's 39.7. This demonstrates that combining explicit reasoning generation with online RL training is a more effective strategy. As a reference, we also include several models from the open-source community that adopt similar discriminative training pipelines. These include model Math-Shepherd-PRM-7B \citep{wang2023math}, based on Mistral-7B, and RLHFlow-Llama3-8B-it \citep{xiong2024implementation}, and Skywork-Qwen2.5-Math-7B-it \citep{skyworkopeno12024}. Notably, the performance of these community-trained models is worse or comparable to that of our reproduced SFT-trained discriminative judge, and they similarly lag far behind our RL-trained \method{} judge. Therefore, we conclude that the stepwise judge benefits from our proposed  recipe of explicit reasoning traces and online reward optimization by RL.

\paragraph{Our RL-trained \method{} judge significantly outperforms existing RL-trained judges} 
Furthermore, we benchmark \method{} against other models trained with online methods like online DPO \citep{xiong2023gibbs, xu2023some} or GRPO \citep{shao2024deepseekmath} (e.g., Eurus-7B, RL-TANGO-7B). Unlike our method, these models are supervised at the 
\textit{trajectory level}, using only the final answer's correctness as a reward signal, denoted by ``Outcome'' in \autoref{tab:process_bench}. The \method{} models, which are trained on explicit step-level signals from Monte-Carlo estimation, also surpass these baselines by a large margin. For example, our best 7B model scores 61.9 while RL-TANGO scores 43.9. This result strongly suggests that direct, step-level supervision provides a much richer and more effective learning signal than a sparse, outcome-only reward.

\paragraph{Test-time compute scaling via majority voting} Since \method{} performs evaluation via CoT reasoning, a natural extension is to generate multiple judgments and use majority voting to decide the final judgment. We apply majority voting with the Qwen2.5-7B-chunk model, and the results are presented in \autoref{tab:process_bench}. We can see that the majority voting shows consistent improvements in the ProcessBench score across various labeling methods. However, the overall gain from majority voting is modest compared to what is often observed in standard mathematical reasoning tasks. We hypothesize this is because our evaluation at each step is binary (correct vs. incorrect), resulting in a much narrower output space than the richer answer spaces typical in math reasoning. In such broader tasks, majority voting is more effective at reducing noise and improving robustness. In contrast, the binary nature of our judgments limits the potential benefit from aggregating multiple outputs.

\begin{table*}[t]
    \centering \scriptsize
           \caption{Ablation study results on ProcessBench. The results show that both the generative CoT reasoning and RL components of our \method{} method are important for overall results. 
           } 
\vspace{10pt} \label{tab:ablation}
 \resizebox{\textwidth}{!}{
    \setlength{\tabcolsep}{4pt}
    \begin{tabular}{l|ccccc}
    \toprule
    \bf Method & \bf GSM8K & \bf MATH & \bf Olympiad & \bf Omni-MATH & \bf Avg $\uparrow$\\
    \midrule 
    \hspace{-1mm}\textit{Qwen2.5-1.5B-chunk} & & & & & \\
    Discriminative + SFT (Baseline) & 32.1 & 32.0 & 14.2 & 18.0 & 24.1 \\
    \textbf{\method{} (Generative Reasoning + RL)} & \textbf{46.9} & \textbf{43.4} & \textbf{26.3} & 28.4 & \textbf{36.2} \\
    ~~ -- Ablate RL (use RS-FT) & 32.8 & 23.9 & 16.3 & 19.6 & 23.1 \\
~~ -- Ablate CoT (use Discriminative format + RL) & 42.0 & 43.2 & 23.6 & \textbf{28.7} & 34.3 \\     \midrule
    \hspace{-1mm}\textit{Qwen2.5-7B-chunk} \\
     Discriminative + SFT (Baseline) &  48.6 & 46.9 & 21.9 & 25.4 & 35.7\\
       \textbf{\method{} (Generative Reasoning + RL)} &  \textbf{72.6} & \textbf{67.2} & \textbf{52.3} & \textbf{49.8} & \textbf{60.5}\\
~~ -- Ablate CoT (use Discriminative format + RL) & 58.7 & 49.4 & 40.8 & 42.7 & 47.9 \\     
~~ -- Ablate Prompt Balancing (Generative Reasoning + RL) & 58.8 & 54.8 & 41.0 & 36.9 & 47.9 \\     
    \bottomrule
    \end{tabular}
    }
    \end{table*}

\subsection{Analysis of the Performance Gap}  \label{sec:understand_performance}

To better understand the source of the performance gap observed on ProcessBench, we isolate the contribution of each component by comparing \method{} against three specialized baselines:
\begin{itemize}
    \item \textbf{Ablate RL}. A \method{} judge trained with rejection sampling fine-tuning (RS-FT, offline): We fine-tune the base model on a static dataset created via rejection sampling, a common offline approach. This isolates the effect of using the CoT format without online RL.
    \item \textbf{Ablate CoT}. A discriminative judge with RL (online): This baseline uses our full RL pipeline but trains the model to output a verdict token directly, without a CoT explanation first. This isolates the effect of online RL without the generative reasoning component.
    \item \textbf{Ablate prompt dataset balancing.} A \method{} judge trained with our full RL pipeline but without prompt dataset balancing. For instance, for \texttt{Rel-Ratio} with a threshold of $0.8$, the proportion of positive samples now is 56.5\% instead of 50\%.
\end{itemize}

\begin{figure*}[t]
    \centering
    \includegraphics[width=0.48\textwidth]{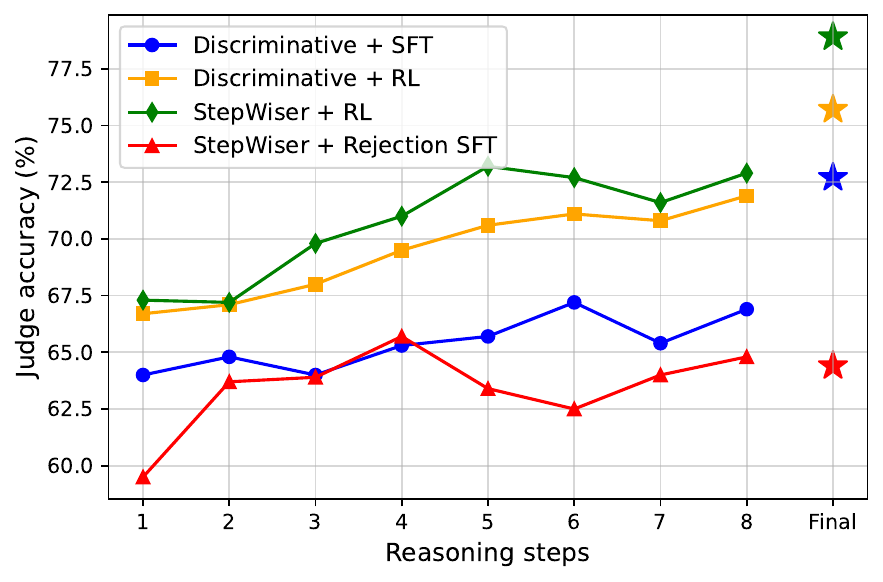}
    \hfill
    \includegraphics[width=0.48\textwidth]{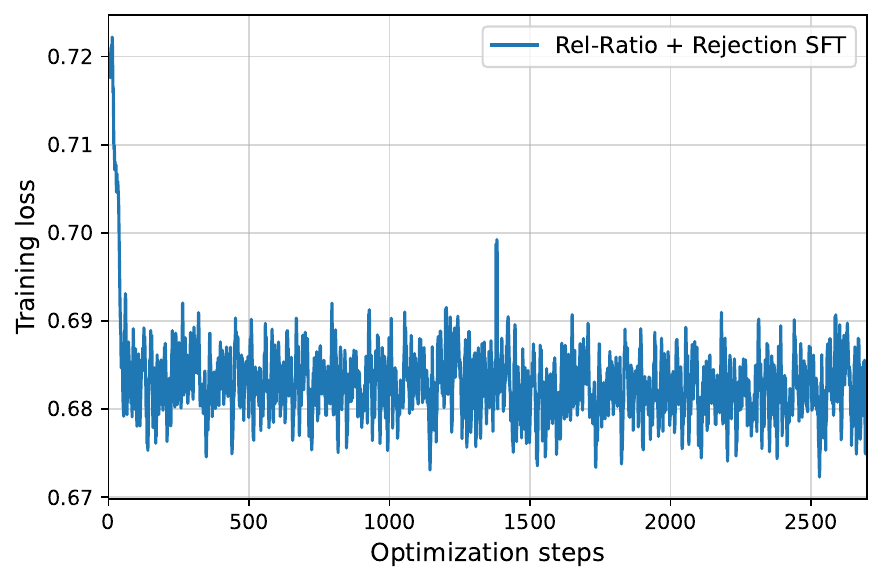}
    \caption{\method{} ablation results. \textbf{Left:} Test stepwise accuracy of various stepwise judge setups. Both generative CoT and RL training are important for the best stepwise judge.  Here we plot the results of \texttt{Rel-Ratio} using Qwen2.5-1.5B-chunk, other results are presented in the Appendix for completeness (see~\autoref{fig:step_accuracy2}). 
    \textbf{Right:} the training loss of rejection sampling fine-tuning, which saturates quickly.}
        \label{fig:step_accuracy} 
\end{figure*}

We evaluate performance using both the ProcessBench score and the in-distribution classification accuracy.  For this study, we focus on the \texttt{Rel-Ratio} signal, where additional results with other labeling approaches are deferred to Appendix since the general trend is similar.  The results, presented in Table~\ref{tab:ablation} and Figure~\ref{fig:step_accuracy}, clearly demonstrate that removing any of theses components leads to a significant drop in performance.

\paragraph{Online RL contributes to the performance improvement} The importance of online learning is evident when comparing our full \method{} model to the RS-FT baseline. On  ProcessBench using Qwen2.5-1.5B-chunk, the RS-FT model achieves an average score of only 23.1, which is substantially lower than  \method{}'s score of 36.2 and is even worse than the standard discriminative SFT baseline (24.1). To understand this phenomenon, we plot the loss curve of rejection sampling fine-tuning in~\autoref{fig:step_accuracy} (right). We notice that its training loss on a large, static dataset plateaus quickly. This trend is consistent across other learning signals and the larger 7B model, indicating that offline methods are insufficient to capture the complexity of CoT reasoning and reward modeling, making online RL a critical component.

\paragraph{\method{} judge with CoT leverages intrinsic reasoning ability to obtain better evaluation} The benefit of the generative CoT format is illustrated by the ``Ablate CoT'' baseline. With the Qwen2.5-1.5B-chunk model, augmenting a discriminative-style judge with RL boosts the ProcessBench score from 24.1 (SFT) to 34.3 (RL), but it still falls short of the \method{} model's 36.2. Moreover, the in-distribution accuracy results in~\autoref{fig:step_accuracy} (left) show that the \method{} model with CoT reasoning achieves higher accuracy on the held-out data. This suggests that generating explicit rationales provides a more expressive and informative structure for learning and modeling the stepwise reward signal. The gap between the generative CoT model and the discriminative model becomes much larger with the stronger Qwen2.5-7B-chunk. Specifically, the generative \method{} model reaches an average score of 60.5, while the discriminative model only achieves 47.9. This is because we are leveraging the intrinsic reasoning ability of the base model through CoT in the judgment so the stronger model offers more advantages.

\paragraph{Prompt dataset balancing stabilizes training and mitigates overfitting}  The practice of balancing the prompt dataset is also crucial for robust performance. Our ablation study on the Qwen2.5-7B-chunk model shows that removing this balancing step causes a substantial performance drop, with the average ProcessBench score dropping from 60.5 to 47.9. A deeper analysis reveals that while both the ``Ablate CoT'' ablation and the lack of dataset balancing hurt performance, their underlying failure modes are different. The ``Ablate CoT'' model suffers from a general decline in its ability to recognize correct and incorrect steps. In contrast, without balancing, the prompt dataset is heavily biased towards positive examples. This trains the model to be overly optimistic, developing a strong bias towards predicting any given step as correct. This bias is particularly enhanced during online training, which eventually leads to training instability and model collapse. A detailed analysis of this phenomenon is provided in the Appendix~\ref{appendix:ablation_cot_balancing}.


\subsection{Using the \method{} judge to obtain better solutions}

In this section, we evaluate the practical utility of our RL-trained \method{} judge in two common applications: guiding an LLM's reasoning process at inference time and selecting high-quality data for subsequent fine-tuning. 
\begin{enumerate}
    \item \textbf{Inference-Time Search via Chunk-Reset Reasoning.}
    To leverage the \method{} judge for improved reasoning, we employ an inference-time search strategy. 
    The base policy model generates a solution ``chunk-by-chunk''. After each chunk is produced, our \method{} judge evaluates it. If the chunk is considered to be good, it is accepted, and the model proceeds to the next step. If it is rejected, the flawed chunk is discarded, and the policy model re-generates a new one from the same point (up to 5 attempts). This allows the model to self-correct and explore alternative reasoning paths without committing to an early mistake, enhancing the final solution's quality. Meanwhile, this reasoning paradigm allows for scaling sequential compute (i.e., the compute used to extend a single inference trajectory with additional steps, rather than running many independent trajectories in parallel), while the overall number of accepted tokens remains similar. 
    \item \textbf{Training Data Selection via Stepwise Rejection Sampling Fine-tuning.} Rejection Sampling Fine-tuning \citep{dong2023raft, touvron2023llama} is a standard technique to improve a base policy by fine-tuning it on its own best outputs. The high-level intuition is that, when the models are allowed to generate $N$ responses per prompt, the pass@N is usually much higher than a randomly selected one (pass@1). Therefore, we can use a proxy reward model to select a training set from the self-generated responses, and train on this set to improve the base policy. However, for mathematical reasoning, verifying only the final answer provides a coarse, binary signal (correct/incorrect) that struggles to differentiate between multiple correct solutions. Here we use \method{}  to compute the scores for each individual reasoning chunk, and use the average score as a proxy to pick the best response from the correct ones.
\end{enumerate}

\begin{table*}[t]
    \centering \footnotesize
        \caption{{\bf Inference time search via Chunk-Reset Reasoning.} 
        We report results with both Qwen2.5-1.5B-chunk and Qwen2.5-7B-chunk, using them as both the response generators and the initialization checkpoints for the \method{} judge. 
We see clear improvements using \method{} across both model sizes, with similar accepted responses lengths (on MATH500).
        Rejected length is the number of tokens in removed chunks during inference time search.
        } \vspace{10pt} \tiny \label{tab:chunk_reset_reasoning}
        \resizebox{\textwidth}{!}{ 
    \setlength{\tabcolsep}{4pt}
    \begin{tabular}{ll|ccc|cc}
    \toprule
         &  \bf Learning &          &  \bf NuminaMath        &  & \bf Accepted & \bf Rejected \\
 \bf Method & \bf signal  & \bf MATH500 & \bf Heldout-1K & \bf Avg $\uparrow$ & \bf length & \bf length \\
    \midrule 
\textit{Qwen2.5-1.5B-chunk} & - & 44.7 & 17.6 & 31.2 & 616.0 & \phantom{000}0.0 \\ 
~~Discriminative + SFT & \texttt{Abs-Q} & 47.7 & 19.1 & 33.4 &  625.2 & \phantom{0}218.7 \\ 
 ~~Discriminative + SFT &  \texttt{Rel-Effective} & 47.4 & 19.6 & 33.5 & 612.7 & \phantom{0}302.4 \\ 
 ~~Discriminative + SFT &  \texttt{Rel-Ratio} & 50.4 & 20.0 & 35.2 & 596.0 & \phantom{0}475.8 \\ 
 ~~Generative CoT + RL (\method{}) &  \texttt{Abs-Q} & 51.4 & 19.8 & 35.6 & 599.1 & 1069.2 \\ 
 ~~Generative CoT + RL (\method{}) &  \texttt{Rel-Effective} & 52.1 & 21.2 & 36.7 & 602.0 & \phantom{0}947.4 \\ 
~~Generative CoT + RL (\method{}) &  \texttt{Rel-Ratio} & 51.9 & 21.8 & 36.9 & 596.4 & \phantom{0}884.7 \\ \midrule 
\textit{Qwen2.5-7B-chunk} &  - & 73.3 & 41.5  & 57.4 & 609.5 & \phantom{000}0.0 \\ 
~~Discriminative + SFT & \texttt{Abs-Q} & 74.8 & 44.4 & 59.6 & 654.0 & \phantom{0}168.2 \\ 
 ~~Discriminative + SFT &  \texttt{Rel-Effective} & 76.9 & 46.1 & 61.5 & 654.6 & \phantom{0}186.5 \\ 
 ~~Discriminative + SFT &  \texttt{Rel-Ratio} & 76.7 & 45.8 & 61.3 & 641.4 & \phantom{0}219.7 \\ 
 ~~Generative CoT + RL (\method{}) & \texttt{Abs-Q} & 77.5 & 46.3 & 61.9 & 658.5 & \phantom{0}345.7 \\ 
   ~~Generative CoT + RL (\method{}) & \texttt{Rel-Effective} &  78.3 & \bf 48.1 & 63.2 &  660.8 & \phantom{0}425.8 \\ 
   ~~Generative CoT + RL (\method{}) & \texttt{Rel-Ratio} & \bf 79.0 & 47.5 & \bf 63.3 & 653.0 & \phantom{0}295.4 \\ 
    \bottomrule
    \end{tabular}
    }
    \end{table*}

\begin{table*}[t]
    \centering \footnotesize
        \caption{{\bf Data selection via Stepwise Rejection Sampling Fine-Tuning.} Our \method{} judge trained with RL provides better quality training data, as measured by final average test performance. The evaluation is with greedy decoding and a maximal generated length of 8192.} \vspace{10pt} \label{tab:rejection_sampling}
    \begin{tabular}{ll|ccc}
    \toprule
 \bf Method & \textbf{Learning signal} & \bf  MATH500 & \bf NM-Heldout-1K & \bf Average $\uparrow$ \\
    \midrule 
\textit{Qwen2.5-7B-chunk (greedy)} &  - & 75.6 & 44.6 & 60.1 \\ \midrule
Outcome-based selection & - & 76.6 & 45.2 & 60.9 \\
Discriminative + SFT & \texttt{Abs-Q} & 78.4 & 45.3 & 61.8  \\
 Discriminative + SFT &  \texttt{Rel-Effective} & 78.2 & 45.2 & 61.7 \\
 Discriminative + SFT &  \texttt{Rel-Ratio} & 78.2 & 45.7 & 61.9 \\
 Generative CoT + RL (\method{}) & \texttt{Abs-Q} & 79.0 & 46.1 & 62.5\\ 
  Generative CoT + RL (\method{}) & \texttt{Rel-Effective} & \textbf{79.4} & 46.7 & \textbf{63.0}\\
   Generative CoT + RL (\method{}) & \texttt{Rel-Ratio} & 79.0 & \textbf{46.8} & 62.9 \\
    \bottomrule
    \end{tabular}
    \end{table*}

The results for inference-time search and data selection are presented in Table~\ref{tab:chunk_reset_reasoning} and Table~\ref{tab:rejection_sampling}, respectively. We summarize the key findings below.
\paragraph{Consistent performance improvements in both setups} In both applications, using our \method{} judge leads to superior outcomes. For inference time search  (Table~\ref{tab:chunk_reset_reasoning}), with the \texttt{Rel-Ratio} learning signal, our approach guides the 1.5B model to achieve an average accuracy of 36.9\%, a significant improvement over the 31.2\% of the base model. We also see clear trends of our \method{} model being superior to the disciminative models across all learning signals. This trend holds for the 7B model, demonstrating the scalability of our method. Similarly, when used for data selection (Table~\ref{tab:rejection_sampling}), models fine-tuned on data selected by the \method{} judge achieve the highest performance (63.0\%), surpassing the original base model (60.1\%), as well as data selected by a discriminative judge (61.9\%) or outcome-based selection (60.9\%).

\paragraph{Superior reasoning error detection using our approach} The ``Accepted Length'' column in Table~\ref{tab:chunk_reset_reasoning} shows that the final solutions from \method{} are of a similar length to the baselines. However, the ``Rejected Length'' column, which corresponds to chunks that judged to be flawed, indicates that our model generates more total tokens to arrive at its final answer. We interpret this as direct evidence of \method{}'s superior ability to identify incorrect or unproductive steps. This triggers the reset mechanism more effectively, forcing the model to discard flawed reasoning and find a better path. This is also consistent with the higher accuracy of identifying the error step on ProcessBench (see also Table~\ref{tab:full_process_bench} in the appendix).

\paragraph{Relative signals are more effective for assigning chunk labels} 
We also observe that across both tables, training signals that reward relative progress (\texttt{Rel-Effective, Rel-Ratio}) consistently yield better judges than a signal that only measures a step's absolute quality (\texttt{Abs-Q}). This appears to be consistent across all results in the paper, for all model sizes and evaluation setups.
For example, in inference time search, the \texttt{Rel-Effective} judge achieves a 64.3\% average accuracy, outperforming the \texttt{Abs-Q} judge (61.9\%). This pattern is also confirmed in data selection, where the \texttt{Rel-Effective} judge produces the best fine-tuning dataset, leading to a final model with 63.0\% accuracy, again surpassing the \texttt{Abs-Q} judge ({62.5\%}).

\section{Conclusion}

Reasoning models that 
output internal thought tokens before a final response have proven to outperform non-reasoning models.
In this paper we have shown that further improvements can be found by making models  {\bf\em reason about the reasoning decisions} made within those internal thoughts. 
We provide a recipe to: (1) segment reasoning into chunks-of-thought;
(2) assign rewards to chunks via relative outcomes of rollouts; and (3) train a judge model to reason about the quality of CoT chunks via reinforcement learning (RL). 

Our  stepwise generative judge \method{}  
is shown to be superior to existing methods on ProcessBench, 
to provide improved inference time search, and better training time rewards for building better response models.
We show that both the use of reasoning during judgment, and training with RL in order to reason about reasoning, are important components to achieve this performance.

\bibliography{iclr2025_conference}
\bibliographystyle{iclr2025_conference}

\appendix

\section{Additional Experiment Details and Results} \label{appendix:early_exp}

\subsection{Discriminative Judge Training and RL Training} \label{appendix:rl_ablation}
\begin{figure*}[htp]
    \centering
    \includegraphics[width=0.48\textwidth]{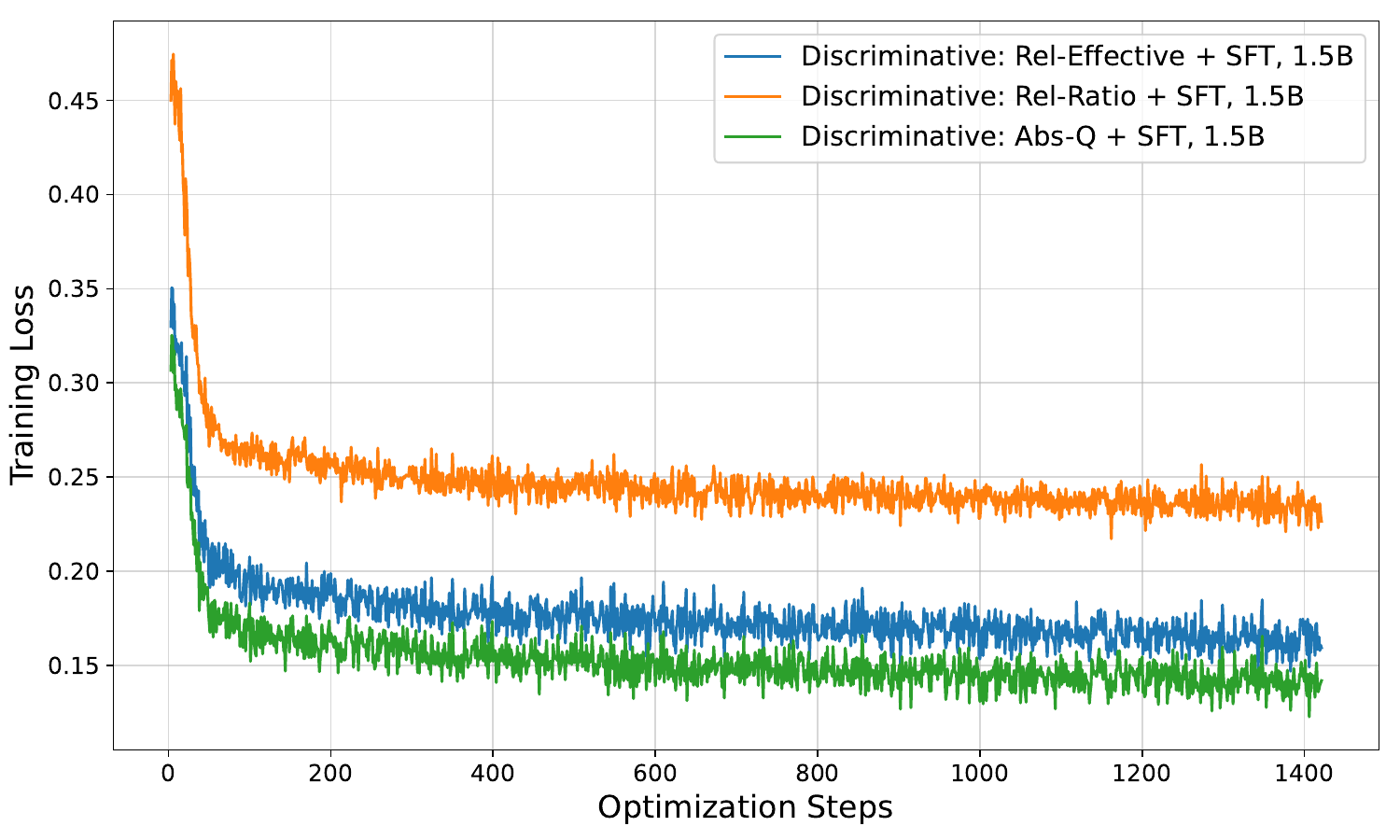}
    \hfill
    \includegraphics[width=0.48\textwidth]{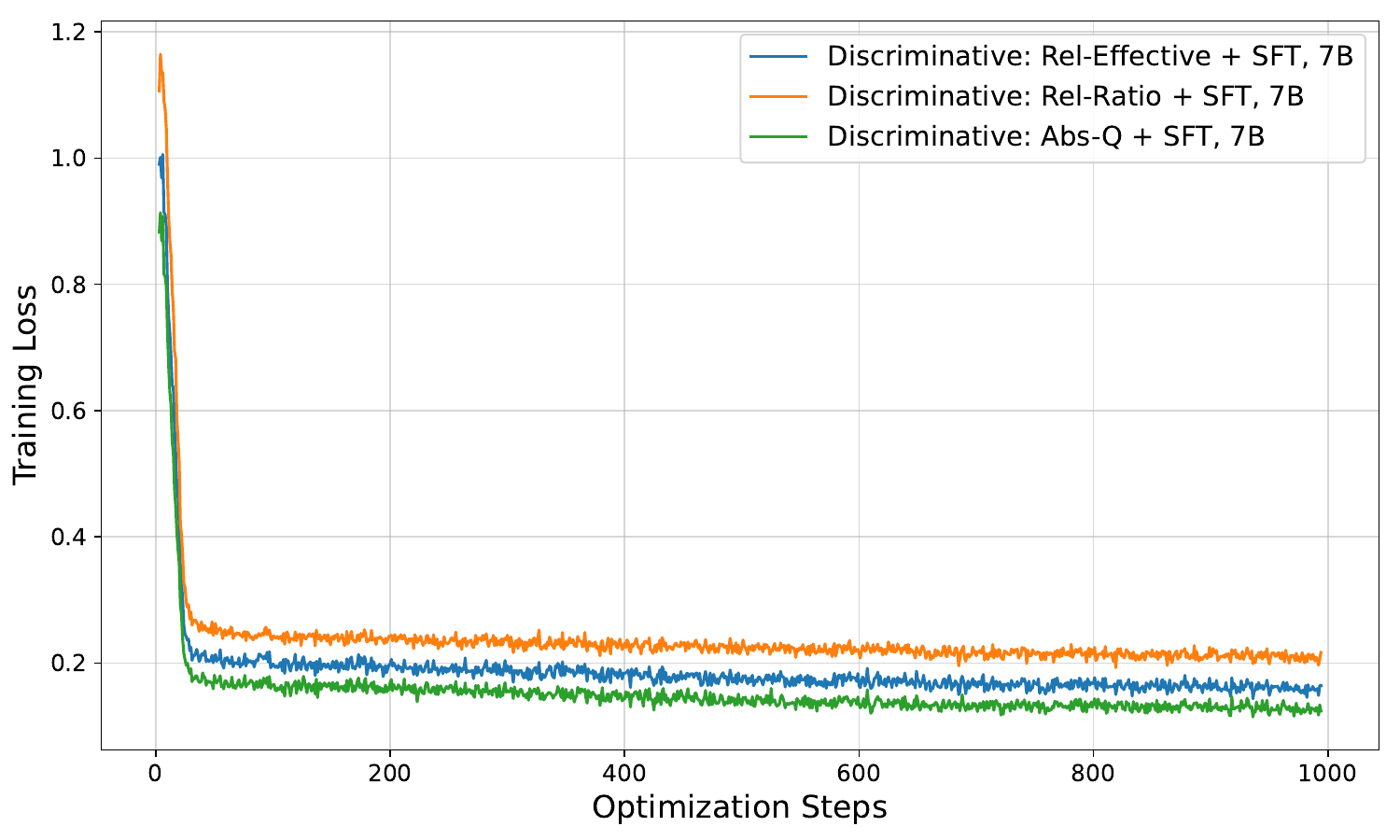}
    \caption{The training loss curves of discriminative stepwise judge under different learning signals. Left: 1.5B model, Right: 7B model.}
        \label{fig:loss_curve_scalar} 
\end{figure*}

\paragraph{Discriminative stepwise judge training}
We follow \citet{xiong2024implementation} to formulate the discriminative stepwise judge as a multi-turn conversation task. Specifically, in every user turn, we provide a single step of reasoning, while in the next assistant turn, the model will decode either ``+'' or ``-'' token to indicate its judgment. 
For training, we use standard SFT code. The data is packed into a block with length 8192 tokens. We use a learning rate of $1e-5$, a global batch size of $32$. We also mask out the user turn's loss. We present the representative training loss curves in Figure~\ref{fig:loss_curve_scalar}.

\paragraph{Hyperparameter search for stepwise labeling}
We conduct hyperparameter tuning for the learning signals labeling. We mainly search by training discriminative models and SFT training, as this is more computationally efficient than full RL training. For \texttt{Rel-Ratio}, we search over $\gamma \in \{0.6, 0.7, 0.8, 1.0, 1.2\}$, and for \texttt{Rel-Effective}, we search over $\alpha \in \{0.2, 0.4, 0.6, 0.8, 1.0\}$ with $\mu$ set as the best-of-4 policy induced by the base policy. For Qwen2.5-1.5B-chunk, we choose $\gamma = 0.8$ and $\alpha = 0.4$, while for Qwen2.5-7B-chunk, we use $\gamma = 0.7$ and $\alpha = 0.8$.

\begin{figure*}[htp]
    \centering
    \includegraphics[width=0.48\textwidth]{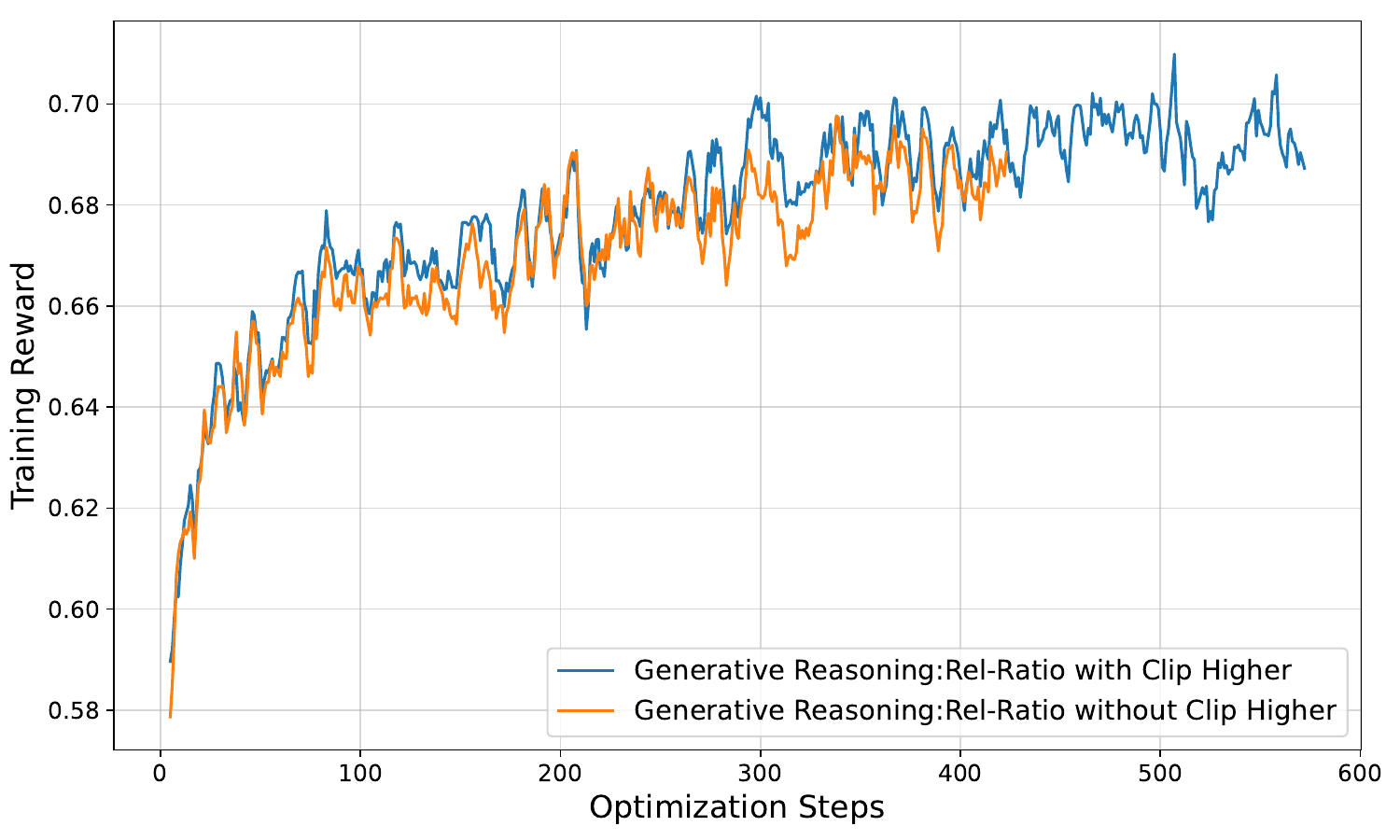}
    \includegraphics[width=0.48\textwidth]{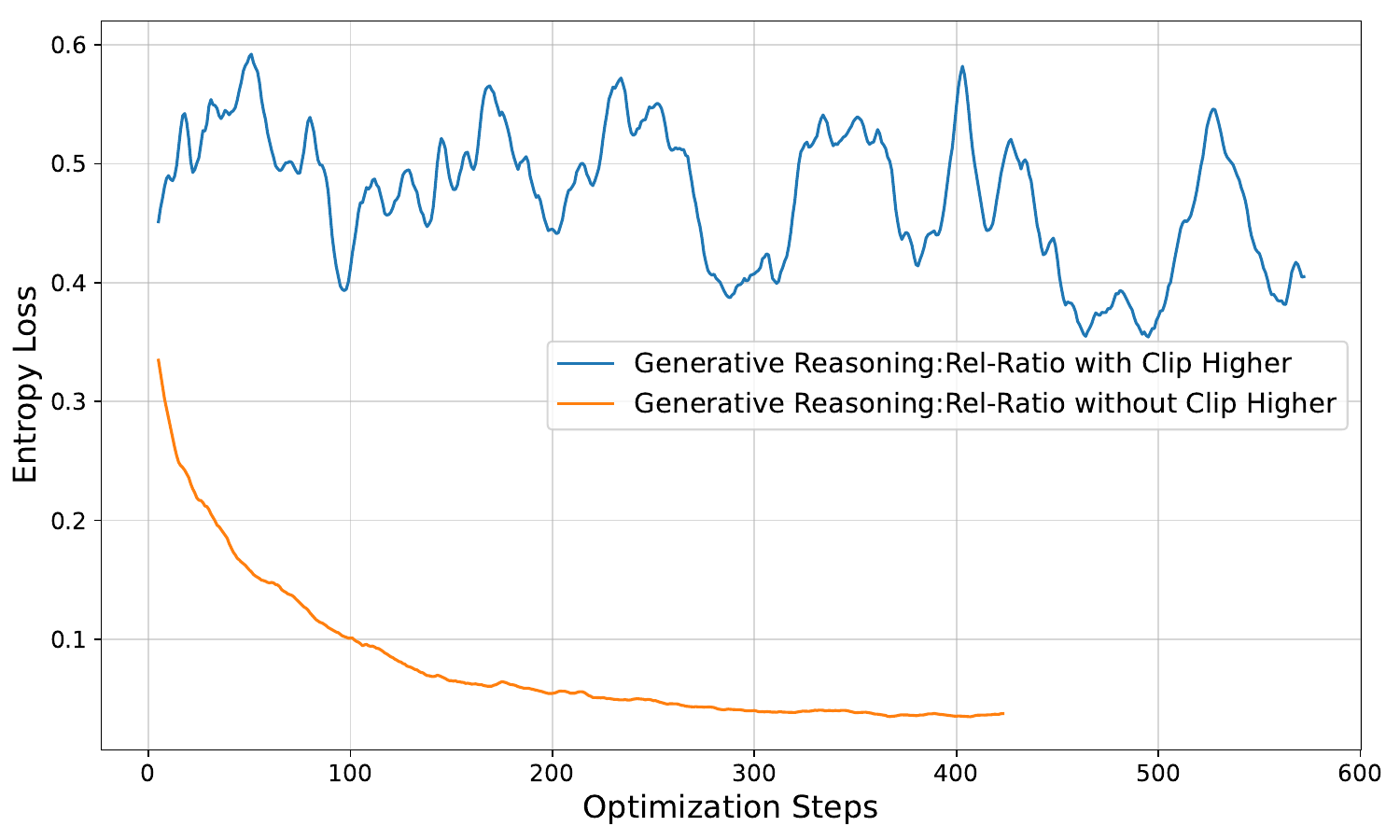}
    \includegraphics[width=0.48\textwidth]{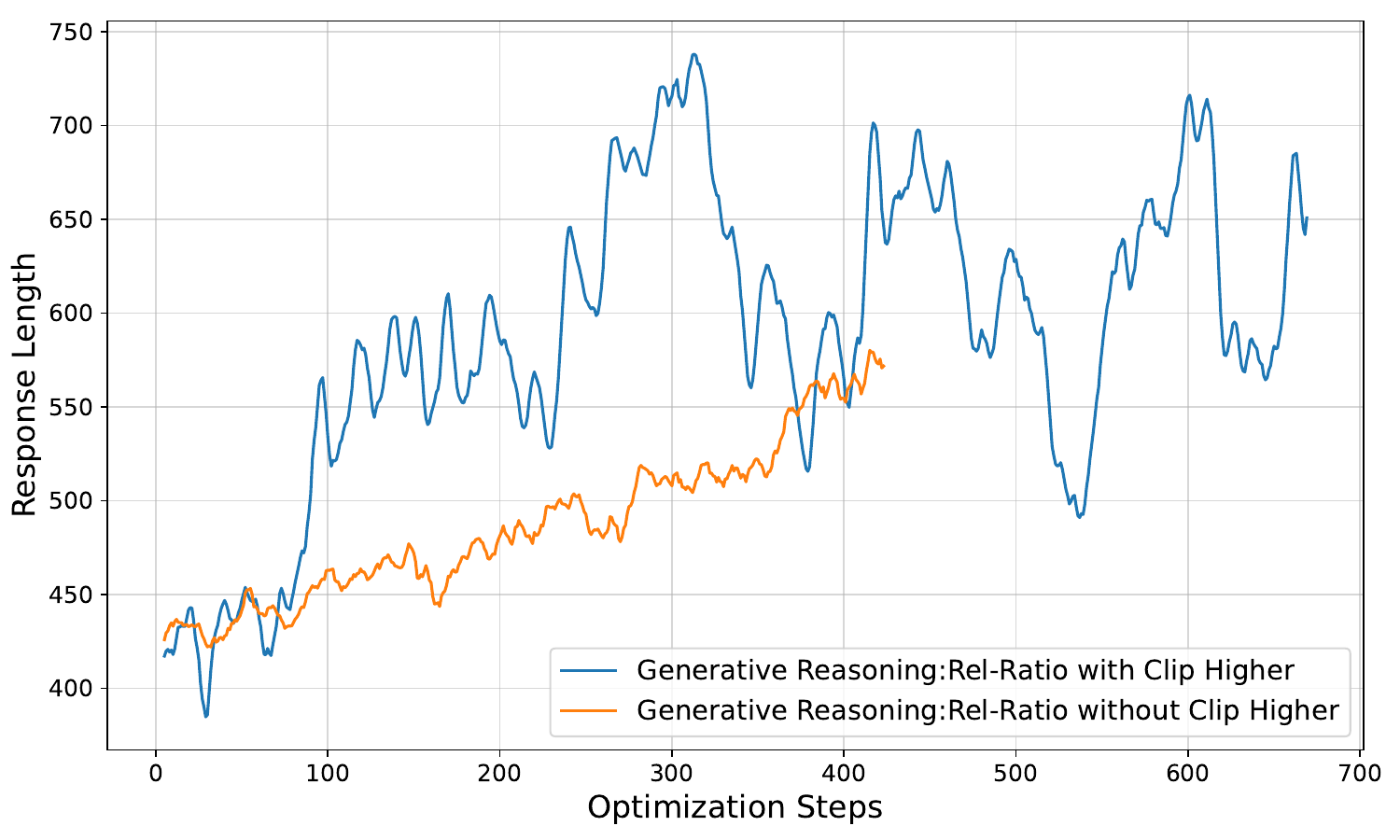}
    \caption{A representative example of training reward, entropy loss, and response length with and without clip higher technique. The model is Qwen2.5-7B-chunk and the learning signal is \texttt{Rel-Ratio} with threshold 0.7.}
        \label{fig:entropy_and_reward} 
\end{figure*}

\paragraph{Learning curves of RL training} In \autoref{fig:entropy_and_reward}, we present a representative example of training reward, entropy loss, and response length with and without clip higher technique. The model is Qwen2.5-7B-chunk and the learning signal is \texttt{Rel-Ratio} with threshold 0.7. We can see that clip higher helps to encourage exploration and leads to a higher training curve.

\subsection{Additional Result on Classification Accuracy}
\begin{figure*}[htp]
    \centering
    \includegraphics[width=0.48\textwidth]{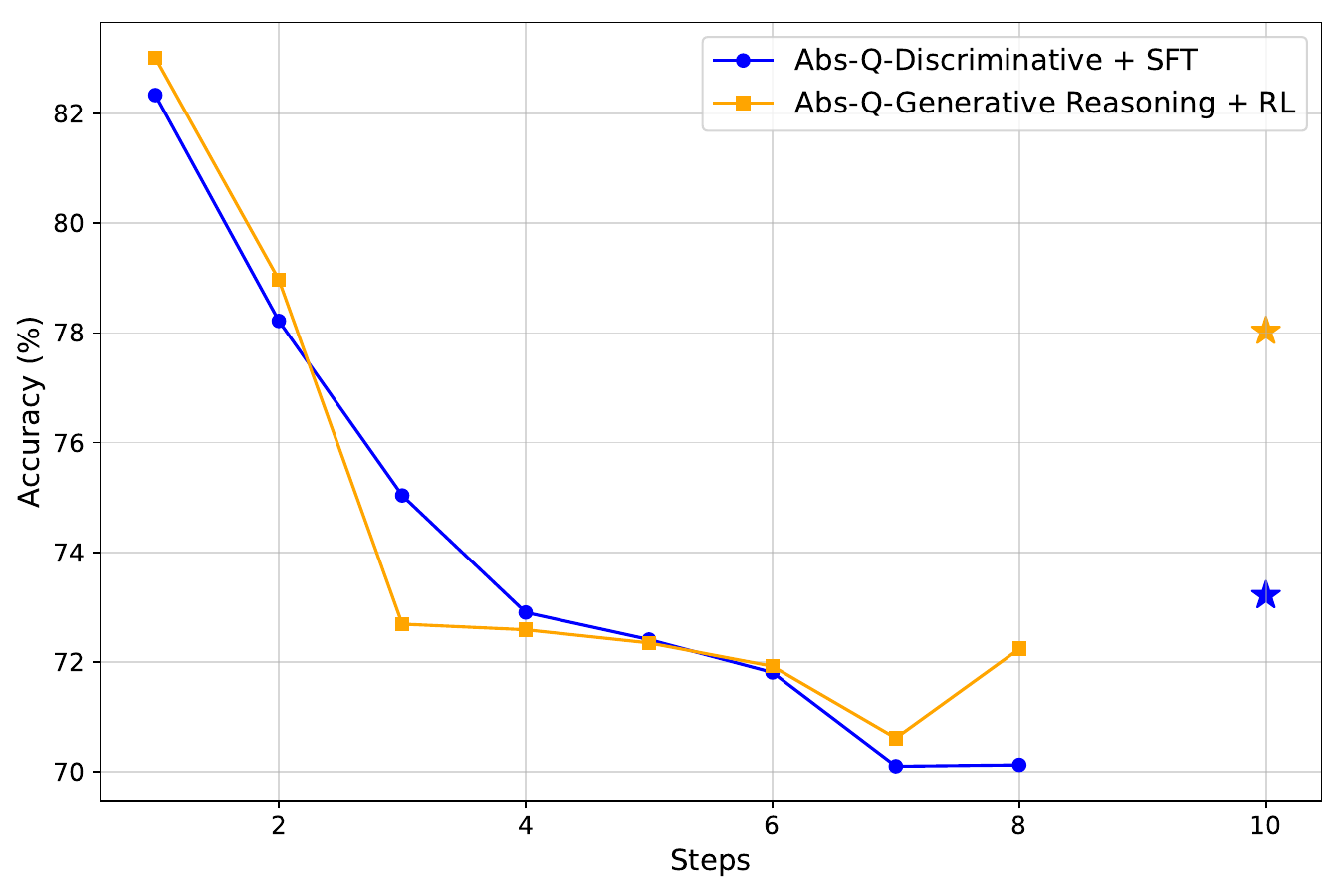}
    \includegraphics[width=0.48\textwidth]{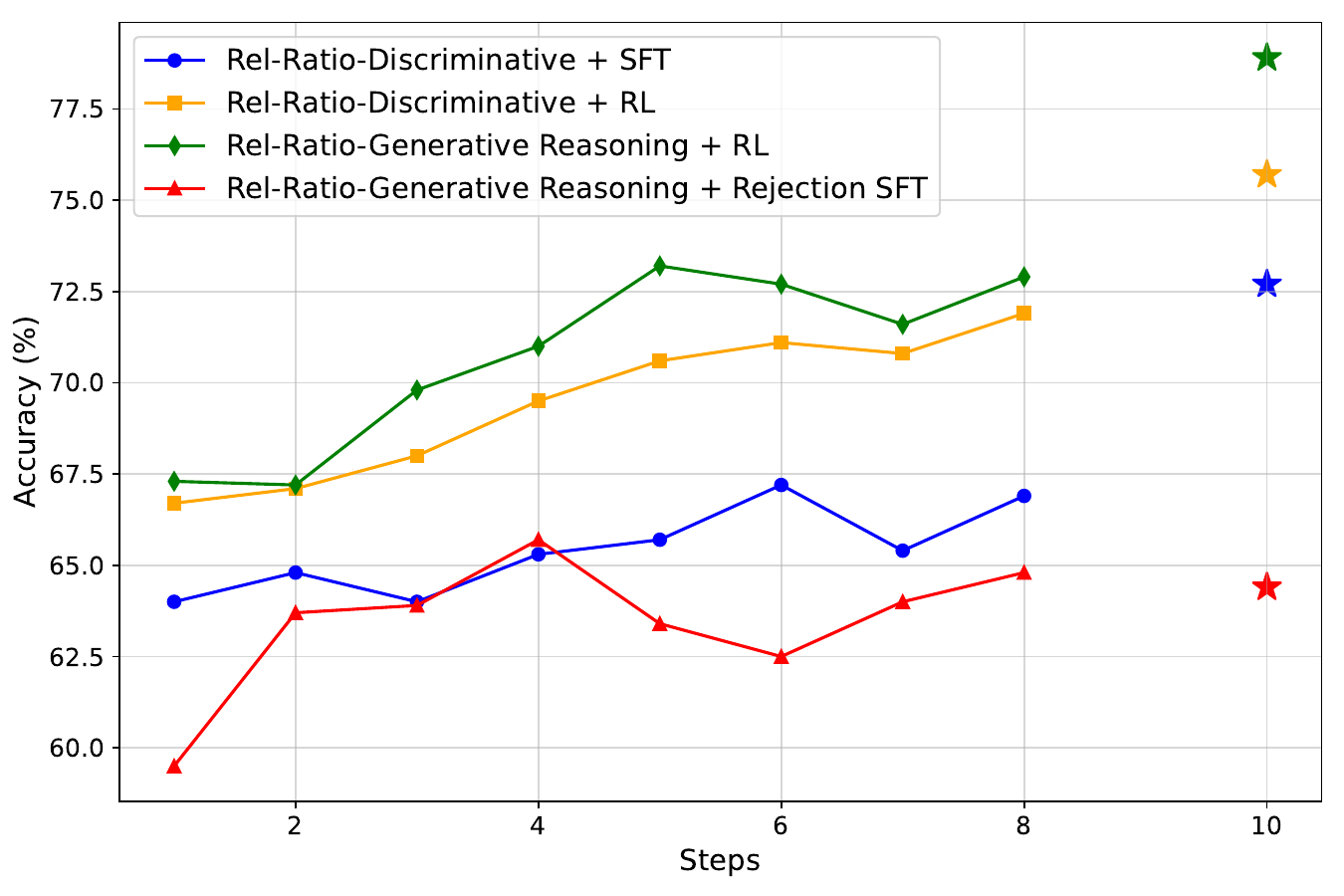}
    \includegraphics[width=0.48\textwidth]{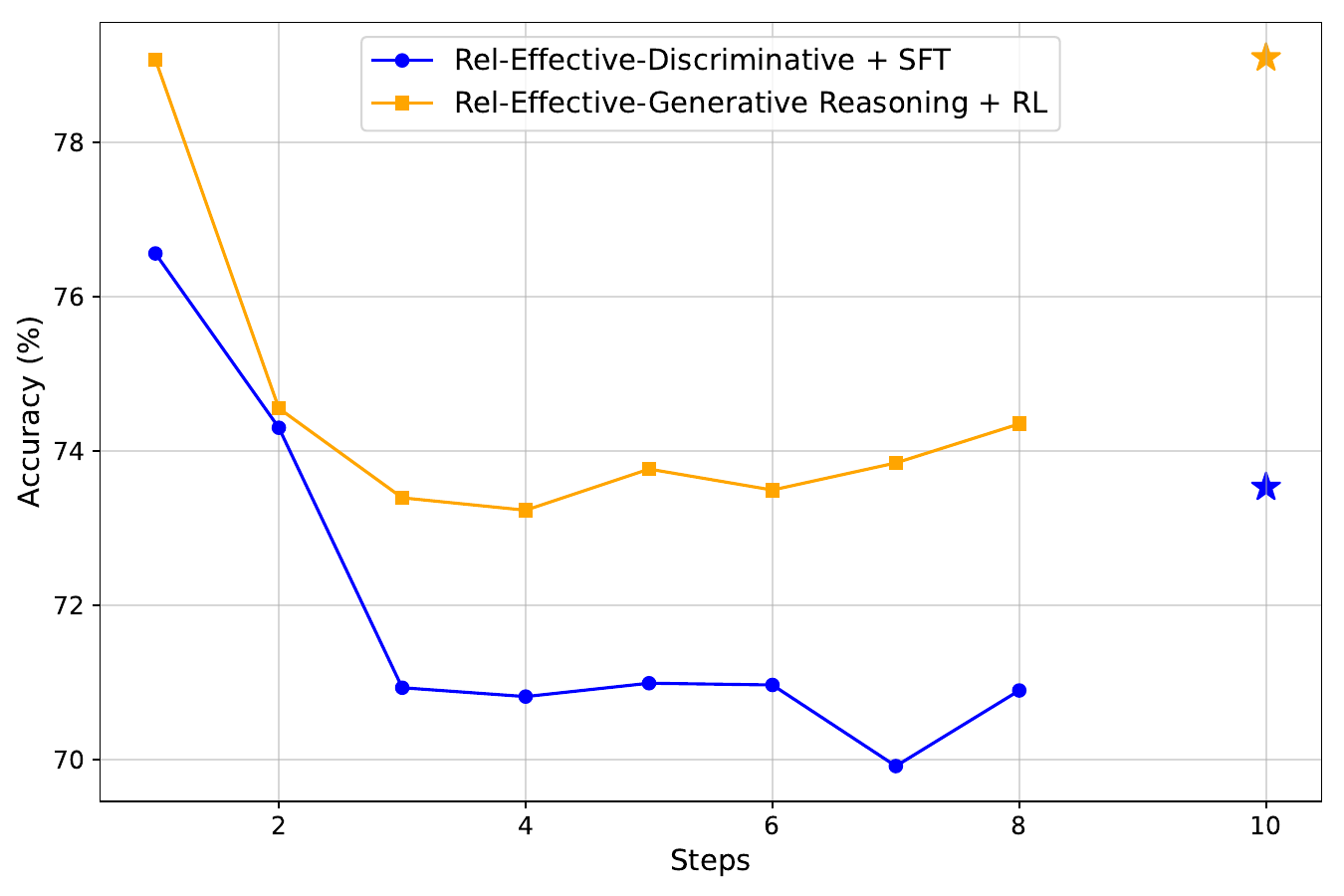}
    \caption{The test stepwise accuracy of different stepwise judges. From left to right, we plot the results of \texttt{Abs-Q, Rel-Ratio, Rel-Effective}, respectively. The stars at step 10 represent the accuracy of recognizing the final answer.}
        \label{fig:step_accuracy2} 
\end{figure*}

We plot the stepwise classification accuracy under different learning signals in Figure~\ref{fig:step_accuracy2}. For \texttt{Rel-Ratio} and \texttt{Rel-Effective}, we observe that the RL-trained generative \method{} judge achieves significantly higher test accuracy on both intermediate steps and final answer correctness. This suggests that the additional reasoning steps enable a more expressive model class that better fits the learning signal. In contrast, for \texttt{Abs-Q}, the performance gap between the discriminative and generative judges is relatively small. We suspect this is because the \texttt{Abs-Q} dataset has a high proportion (70.2\%) of positive samples. To stabilize RL training, we downsample the positive class, which may introduce distribution shift and affect test classification accuracy. Nevertheless, our \method{} model still achieves substantially higher accuracy in evaluating final answer correctness.

\subsection{Additional Result on Self-segmentation Fine-tuning}
\label{appendix:chunking}

\begin{table*}[h]
    \centering \footnotesize
        \caption{The main ablation results on self-segmentation fine-tuning an chunking.} \vspace{10pt} \label{tab:chunking_stat}
    \setlength{\tabcolsep}{4pt}
    \begin{tabular}{ll|cccccc}
    \toprule
 Method & \textbf{Learning signal} & \# Steps &  GSM8K & MATH & Olympiad & Omni-MATH & Ave \\
    \midrule 
Split by \textbackslash n\textbackslash n  & \texttt{Abs-Q} + SFT & 5457820 & 33.7 & 37.1 & 20.2 & 18.9 & 27.5 \\
Split by \textbackslash n\textbackslash n  & \texttt{Abs-Q} + RL & - & 46.3 & 38.4 & 19.0 & 25.8 & 32.4 \\
Split by \textbackslash n\textbackslash n  & \texttt{Rel-Ratio} + SFT & - &  28.3 & 30.9 & 15.5 & 21.0 & 23.9\\ 
Split by \textbackslash n\textbackslash n  & \texttt{Rel-Ratio} + RL & - & 46.3 & 39.1 & 17.3 & 21.3 & 31.0 \\ \midrule
 Self-segmentation & \texttt{Abs-Q} + SFT & 3463520 & 39.3 & 32.1 & 19.3 & 18.9 & 27.2  \\ 
 Self-segmentation & \texttt{Abs-Q} + RL & - & 49.2 & 40.5 & 23.8 & 31.0 & 36.1  \\ 
Self-segmentation & \texttt{Rel-Ratio} + SFT & - & 32.1 & 32.0 & 14.2 & 18.0 & 24.1 \\
Self-segmentation & \texttt{Rel-Ratio} + RL& - & 46.9 & 43.4 & 26.3 & 28.4 & 36.2  \\
    \bottomrule
    \end{tabular}
    \end{table*}

In terms of task performance, the benefits of self-segmentation are most apparent in the context of RL training. While the average scores for SFT are closer (27.5 vs. 27.2 for \texttt{Abs-Q}), models trained with RL show significant improvements. Specifically, the average score for \texttt{Abs-Q} + RL increased from 32.4 to 36.1, and the \texttt{Rel-Ratio} + RL score rose from 31.0 to 36.2. This disparity suggests that the self-segmentation process effectively filters out noisy and less informative steps from the reasoning trajectories. The resulting cleaner signal is highly beneficial for the RL training process, which is more sensitive to data quality. This observation is also consistent with our findings from the prompt filtering and balancing experiments. Conversely, SFT appears more robust to this type of noise, and thus its performance is less impacted. 

\subsection{Additional Ablation Result on CoT and Prompt Dataset Balancing} \label{appendix:ablation_cot_balancing}

\paragraph{Prompt dataset balance technique} In our early experiments, we observed that the prompt sets could be highly imbalanced. For example, using the Qwen2.5-1.5B-chunk model, the proportion of positive samples for \texttt{Rel-Ratio} with a threshold of 0.8 is 56.5\%, while for \texttt{Abs-Q}, it is 70.2\%. Such imbalance may cause model degeneration, as the model can achieve a high score by simply predicting ``correct'' in most cases. To address this, we consider a prompt balance technique that down-samples the majority class. 

\paragraph{Generative judge without CoT} Another key component of our approach is the use of CoT reasoning before producing the final judgment. To assess its impact on performance, we also experiment with prompting the model to generate the final judgment directly, while assigning a format penalty of $-1.0$ whenever the model fails to follow the requirement. Under this setting, the model quickly learns to produce judgments without CoT reasoning in fewer than 20 training steps.

\paragraph{Results}
We present the detailed results of ablation on CoT and prompt dataset balancing in \autoref{tab:full_process_bench}. The results reveal that while the absence of either component degrades the final F1 score, the underlying reasons for the performance drop are fundamentally different.

The removal of CoT appears to weaken the model's overall ability to discriminate between correct and incorrect reasoning steps. This is evidenced by a general decline in accuracy for both the ``Correct'' and ``Error'' classifications across the datasets. In contrast, removing dataset balancing introduces a strong class bias. Without balancing, the model is trained on a dataset where correct steps are overrepresented, causing it to overfit and develop a tendency to classify most steps as correct. This is clearly demonstrated by a sharp increase in accuracy for the ``Correct'' class, coupled with a significant decrease in accuracy for the ``Error'' class. While the model becomes proficient at identifying correct steps, it largely loses its ability to detect errors. This trade-off is ultimately detrimental, as shown by the identical drop in the average F1 score from 60.5 to 47.9 for the Rel-Ratio model.

\section{Template, Example, and Additional Tables}

\begin{table}[!t]
\setlength{\tabcolsep}{4pt}
\renewcommand{\arraystretch}{1.1}
\caption{\small Judge performance on ProcessBench, broken down by four subsets. Each subset reports Error (\%), Correct (\%), and F1 score (\%). The final column is the average F1 across all subsets. We remark that the F1 score here is indeed the harmonic mean of the accuracies on two classes.}
\vspace{-1mm}
\label{tab:full_process_bench}
\small
\begin{center}
\resizebox{\linewidth}{!}{
\setlength{\tabcolsep}{4pt}
\begin{tabular}{ll|ccc|ccc|ccc|ccc|c}
\toprule[1pt]
\textbf{Method} & \textbf{Learning} & 
\multicolumn{3}{c|}{\textbf{GSM8K}} & 
\multicolumn{3}{c|}{\textbf{MATH}} & 
\multicolumn{3}{c|}{\textbf{Olympiad}} & 
\multicolumn{3}{c|}{\textbf{Omni-MATH}} & 
\textbf{Avg. F1} \\
\cmidrule(lr){3-5} \cmidrule(lr){6-8} \cmidrule(lr){9-11} \cmidrule(lr){12-14}
&  {\bf ~~Signal} & Error & Correct & F1 & Error & Correct & F1 & Error & Correct & F1 & Error & Correct & F1 & \\
\midrule
\textit{Qwen2.5-1.5B-chunk} & \\
Discriminative + SFT & \texttt{Abs-Q} & 26.0 & 80.0 & 39.3 & 22.2 & 57.6 & 32.1 & 14.2 & 30.2 & 19.3 & 13.2 & 28.2 & 18.0 & 27.2\\
Discriminative + SFT &  \texttt{Rel-Effect} & 28.5 & 72.0 & 40.8 & 28.6  & 53.0 & 37.2 & 16.4 & 21.8 & 18.7 & 15.8 & 27.6 & 20.1 & 29.2\\
Discriminative + SFT & \texttt{Rel-Ratio} & 22.5 & 56.0 & 32.1 & 26.2 & 41.0 & 32.0 & 14.0 & 14.4 & 14.2 & 15.2 & 22.0 & 18.0 & 24.1 \\
Generative + CoT + RL &  \texttt{Abs-Q} & 42.5 & 58.5 & 49.2 & 36.4 & 45.6 & 40.5 & 31.4 & 19.2 & 23.8 & 32.8 & 29.4 & 31.0 & 36.1 \\
Generative + CoT + RL &  \texttt{Rel-Effect} & 38.5 & 64.5 & 48.2 & 37.8 & 51.6 & 43.6 & 23.2 & 21.0 & 22.1 & 24.0 & 26.8 & 25.3 & 34.8 \\
Generative + CoT + RL &  \texttt{Rel-Ratio}& 35.0 & 71.0 & 46.9 & 37.8 & 50.8 & 43.4 & 27.0 & 25.6 & 26.3 & 28.0 & 28.8 & 28.4 & 36.2 \\
Gen  + RL (no CoT) &  \texttt{Rel-Ratio} & 28.5 & 79.5 & 42.0 & 37.0 & 51.8 & 43.2 & 24.4 & 22.8 & 23.6 & 28.6 & 28.8 & 28.7 & 34.3 \\
Gen + CoT + RL (no Chunk) & \texttt{Rel-Ratio} & 36.0 & 65.5 & 46.5 & 37.0 & 39.8 & 38.4 & 25.4 & 15.2 & 19.0 & 29.4 & 23.0 & 25.8 & 32.4\\
\midrule
\textit{Qwen2.5-7B-chunk} & \\
Discriminative + SFT & \texttt{Abs-Q} & 41.0 & 80.5 & 54.3 & 36.0 & 66.4 & 46.7 & 28.8 & 43.4 & 34.6 & 21.8 & 39.6 & 28.1 & 40.9\\
Discriminative + SFT &  \texttt{Rel-Effect} & 40.5 & 80.0 & 53.8 & 36.8 & 69.6 & 48.1 & 27.0 & 36.2 & 30.93 & 24.6 & 41.0 & 30.8 &38.7 \\
Discriminative + SFT & \texttt{Rel-Ratio} & 37.5 & 78.5 & 50.8 & 36.6 & 63.8 & 46.5 & 24.0 & 35.8 & 28.7 & 24.2 & 35.6 & 28.8 & 38.7\\
Generative + CoT + RL &  \texttt{Abs-Q} & 59.5 & 64.5 & 61.9 & 63.2 & 59.0 & 61.0 & 53.0 & 44.6 & 48.4 & 44.4 & 43.4 & 43.9 & 53.8 \\
Generative + CoT + RL &  \texttt{Rel-Effect} & 70.5 & 74.5 & 72.4 & 69.2 & 67.4 & 68.3 & 61.4 & 48.8 & 54.4 & 54.4 & 50.6 & 52.4 & 61.9\\
Generative + CoT + RL &  \texttt{Rel-Ratio} & 66.5 & 80.0 & 72.6 & 62.6 & 72.6 & 67.2 & 57.2 & 48.2 & 52.3 & 49.4 & 50.2 & 49.8 & 60.5 \\
\midrule
\textit{Qwen2.5-7B-chunk} Ablation & \\
Gen + CoT + RL (no Balancing) &  \texttt{Abs-Q} & 31.5 & 94.0 & 47.2 & 34.0 & 79.6 & 47.7 & 25.0 & 58.0 & 34.9 & 23.2 & 43.2 & 30.2 & 40.0 \\
Gen + CoT + RL (no Balancing) &  \texttt{Rel-Effect} &45.0 & 94.0 & 60.9 & 44.8 & 79.0 & 57.2 & 35.8 & 59.2 & 44.6 & 27.0 & 48.8 & 34.8 & 49.4 \\
Gen + CoT + RL (no Balancing) &  \texttt{Rel-Ratio} & 42.5 & 95.5 & 58.8 & 41.6 & 80.2 & 54.8 & 29.8 & 65.8 & 41.0 & 29.4 & 49.6 & 36.9 & 47.9 \\
Gen + RL (no CoT) &  \texttt{Rel-Ratio} & 45.5 & 82.5 & 58.7 & 37.6 & 72.0 & 49.4 & 36.0 & 47.0 & 40.8 & 40.2 & 45.6 & 42.7 & 47.9 \\
\bottomrule[1pt]
\end{tabular}}
\end{center}
\vspace{-5mm}
\end{table}

\begin{table}
\caption{Prompt template for solver model to produce self-segmented CoTs.}
\small
\begin{tcolorbox}[
    colback=gray!5,
    colframe=gray!60!black,
    title=Prompt Template for Policy Model,
    fonttitle=\bfseries,
    sharp corners,
    boxrule=1pt,
]
You are a helpful assistant designed to solve mathematical problems step-by-step. Your task is to think step-by-step and provide a detailed solution process following a specific format. 

You \textbf{MUST} follow the following format:
\begin{enumerate}[label=\arabic*., topsep=3pt, itemsep=2pt, leftmargin=*]
    \item Enclose each part of your step-by-step reasoning within \texttt{\textless chunk\textgreater} and \texttt{\textless /chunk\textgreater} tags.
    \item After completing the chain-of-thought reasoning, provide the final answer within \texttt{\textbackslash boxed\{\{\}\}}.
\end{enumerate}

Ensure strict compliance with this format for every response.  

\vspace{1em}
\hrule 
\vspace{1em}

\textbf{Problem:} \texttt{\{problem\}}
\label{tab:prompt_chunking_sft}\vspace{5pt}
\end{tcolorbox}
\end{table}

\begin{table*}
    \caption{A representative example illustrating CoT segmentation methods based on a response from Qwen2.5-1.5B-it. Standard splitting by newlines (left) produces too many chunks without a clear logical step. For example, chunk 3 and chunk 4 are logically a whole, but they are separated, resulting in poor independence of each step and the inability to form a complete thinking unit.\label{tab:chunking2} }
\vspace{6pt}
\centering
\small
\begin{tabular}{|m{0.47\textwidth}|m{0.47\textwidth}|}
\hline
\centering\textbf{Split by \texttt{\textbackslash n\textbackslash n}} & \centering\textbf{LLM-based self-segmentation (Chunks-of-Thought)} \tabularnewline
\hline

\textbf{chunk 1} \par
To find the value of \( f(5) \) for the given function $$f(x) = \begin{cases} x-3, & x\geq6 \\ f(f(x+5)), & x<6\end{cases} ,$$ we need to follow the recursive definition of the function for \( x < 6 \).

\vspace{6pt}

\textbf{chunk 2} \par
Since \( 5 < 6 \), we use the second case of the piecewise function: \[ f(5) = f(f(5+5)) = f(f(10)). \]

\vspace{6pt}
\textbf{chunk 3} \par
Now, we need to evaluate \( f(10) \). Since \( 10 \geq 6 \), we use the first case of the piecewise function: \[ f(10) = 10 - 3 = 7. \]

\vspace{6pt}
\textbf{chunk 4} \par
So, we have: \[ f(5) = f(7). \]

\vspace{6pt}
\textbf{chunk 5} \par
Next, we need to evaluate \( f(7) \). Again, since \( 7 \geq 6 \), we use the first case of the piecewise function: \[ f(7) = 7 - 3 = 4. \]

\vspace{6pt}
\textbf{chunk 6} \par
Thus, we have: \[ f(5) = 4. \]

\vspace{6pt}
\textbf{chunk 7} \par
The final answer is: \[ \boxed{4}\].
 &

\textbf{chunk 1 (Initial Expansion)} \par
To find the value of \( f(5) \) for the given function $$ f(x) = \begin{cases} x-3, & x\geq6 \\ f(f(x+5)), & x<6\end{cases} ,$$ we need to follow the recursive definition of the function for \( x < 6 \).

Since \( 5 < 6 \), we use the second case of the piecewise function:
\[ f(5) = f(f(5+5)) = f(f(10)). \]

\vspace{6pt}

\textbf{chunk 2 (Calculate Inner Function and Substitute)} \par
Now, we need to evaluate \( f(10) \). Since \( 10 \geq 6 \), we use the first case of the piecewise function:
\[ f(10) = 10 - 3 = 7. \]
So, we have:
\[ f(5) = f(7). \]

\vspace{6pt}

\textbf{chunk 3 (Final Calculation)} \par
Next, we need to evaluate \( f(7) \). Again, since \( 7 \geq 6 \), we use the first case of the piecewise function:
\[ f(7) = 7 - 3 = 4. \]
Thus, we have:
\[ f(5) = 4. \]

The final answer is:
\[
\boxed{4}
\]
\tabularnewline
\hline
\end{tabular}
\end{table*}

\newpage

\begin{table*}
    \caption{A representative example illustrating CoT segmentation methods based on a response from Qwen2.5-1.5B-it. Standard splitting by newlines (left) produces too many chunks without a clear logical step. \label{tab:chunking4} }
\vspace{6pt}
\centering
\tiny
\begin{tabular}{|m{0.47\textwidth}|m{0.47\textwidth}|}
\hline
\centering\textbf{Split by \texttt{\textbackslash n\textbackslash n}} & \centering\textbf{LLM-based self-segmentation (Chunks-of-Thought)} \tabularnewline
\hline

\textbf{chunk 1} \par
To solve the given problem, we start by expressing \( z \) in its standard form \( z = x + yi \), where \( x \) and \( y \) are real numbers. The given equation is:

\vspace{6pt}

\textbf{chunk 2} \par
\[
\left|\frac{z^2+1}{z+i}\right| + \left|\frac{z^2+4i-3}{z-i+2}\right| = 4.
\]
\vspace{6pt}
\textbf{chunk 3} \par
First, let's express the magnitudes of the complex numbers. For the first term, we have:

\vspace{6pt}
\textbf{chunk 4} \par
\[
\left|\frac{z^2+1}{z+i}\right| = \left| \frac{(x+yi)^2 + 1}{x+yi + i} \right|.
\]
\vspace{6pt}
\textbf{chunk 5} \par
Since \( (x+yi)^2 = x^2 - y^2 + 2xyi \), we get \( z^2 + 1 = x^2 - y^2 + 2xyi + 1 = (x^2 - y^2 + 1) + 2xyi \). Therefore, the magnitude is:

\vspace{6pt}
\textbf{chunk 6} \par
$$
\begin{aligned}
    &\left| (x^2 - y^2 + 1) + 2xyi + i(x + yi) \right| \\&= \left| (x^2 - y^2 + 1 + x + y) + (2xy + x + y)i \right|.
\end{aligned}
$$

\vspace{6pt}
\textbf{chunk 7} \par
This magnitude can be simplified to the square root of the sum of the squares of the real and imaginary parts:

\vspace{6pt}
\textbf{chunk 8} \par
\[
\sqrt{(x^2 - y^2 + 1 + x + y)^2 + (2xy + x + y)^2}.
\]
\vspace{6pt}
\textbf{chunk 9} \par
Similarly, for the second term, we have:

\vspace{6pt}
\textbf{chunk 10} \par
\[
\left|\frac{z^2 + 4i - 3}{z - i + 2}\right| = \left| \frac{(x+yi)^2 + 4i - 3}{x+yi - i + 2} \right|.
\]
\vspace{6pt}
\textbf{chunk 11} \par
Since \( z^2 + 4i - 3 = x^2 - y^2 + 2xyi + 4i - 3 = (x^2 - y^2 - 3) + (2xy + 4)i \), we get: 

\vspace{6pt} 
\textbf{chunk 12} \par
\[
z - i + 2 = x + (y-1) = x + yi - i + 2 = (x + 2) + yi.
\]
\vspace{6pt} \textbf{chunk 13} \par
Using the same logic as above, the magnitude is:
\vspace{6pt} \par
...
\vspace{6pt}

\textbf{chunk 21} \par
\[
\boxed{0}.
\]
 & 
\textbf{chunk 1 (Express the Equation in Terms of Real and Imaginary Parts)} \par
To solve the given problem, we start by expressing \( z \) in its standard form \( z = x + yi \), where \( x \) and \( y \) are real numbers. The given equation is:
\[
\left|\frac{z^2+1}{z+i}\right| + \left|\frac{z^2+4i-3}{z-i+2}\right| = 4.
\]
\vspace{6pt}

\textbf{chunk 2 (Simplify the first Term)} \par

First, let's express the magnitudes of the complex numbers. For the first term, we have:
\[
\left|\frac{z^2+1}{z+i}\right| = \left| \frac{(x+yi)^2 + 1}{x+yi + i} \right|.
\]
Since \( (x+yi)^2 = x^2 - y^2 + 2xyi \), we get \( z^2 + 1 = x^2 - y^2 + 2xyi + 1 = (x^2 - y^2 + 1) + 2xyi \). Therefore, the magnitude is:
$$
\begin{aligned}
    &\left| (x^2 - y^2 + 1) + 2xyi + i(x + yi) \right| \\&= \left| (x^2 - y^2 + 1 + x + y) + (2xy + x + y)i \right|.
\end{aligned}
$$
This magnitude can be simplified to the square root of the sum of the squares of the real and imaginary parts:
\[
\sqrt{(x^2 - y^2 + 1 + x + y)^2 + (2xy + x + y)^2}.
\]

\vspace{6pt}

\textbf{chunk 3 (Simplify the Second Term and Formulate Equations)} \par
Similarly, for the second term, we have:
\[
\left|\frac{z^2 + 4i - 3}{z - i + 2}\right| = \left| \frac{(x+yi)^2 + 4i - 3}{x+yi - i + 2} \right|.
\]
Since \( z^2 + 4i - 3 = x^2 - y^2 + 2xyi + 4i - 3 = (x^2 - y^2 - 3) + (2xy + 4)i \), we get:
\[
z - i + 2 = x + (y-1) = x + yi - i + 2 = (x + 2) + yi.
\]
Using the same logic as above, the magnitude is:
$$
\begin{aligned}
    &\sqrt{\left(x^2 - y^2 - 3 + x + y + 2\right)^2 + \left(2xy + 4 + x + y\right)^2} \\&= \sqrt{\left(x^2 - y^2 - 1 + x + y\right)^2 + \left(2xy + x + y + 4\right)^2}.
\end{aligned}
$$

\vspace{6pt}

\textbf{chunk 4 (Solve the System of Equations to Find z)} \par
Given that these two magnitudes sum to 4, and recognizing that the equality \( a^2 + b^2 = 4 \) indicates that each term must separately be 2 (because the maximum magnitude of any complex number is 2), we derive:

\[
(x^2 - y^2 + 1 + x + y)^2 + (2xy + x + y)^2 = 4,
\]
\[
\left(x^2 - y^2 - 3 + x + y + 2\right)^2 + \left(2xy + 4 + x + y\right)^2 = 4.
\]

This leads to \( x^2 - y^2 + x = 1 \) and \( x^2 - y^2 - x = -1 \). Solving these equations simultaneously, we get \( x = 1 \) and \( y = 0 \), so \( z = 1 \).

\vspace{6pt}

\textbf{chunk 5 (Calculate the Final Minimum Value)} \par
Finally, we need to find the minimum value of \( |z - 1| \). This simplifies to:

\[
|z - 1| = |1 - 1| = 0.
\]

Thus, the minimum value of \( |z - 1| \) is:

\[
\boxed{0}.
\]

\tabularnewline
\hline
\end{tabular}
\end{table*}

\newpage

\begin{table*}
    \caption{A representative example illustrating CoT segmentation methods based on a response from Qwen2.5-1.5B-it. Standard splitting by newlines (left) produces too many chunks without a clear logical step. \label{tab:chunking5} }
\vspace{6pt}
\centering
\tiny
\begin{tabular}{|m{0.47\textwidth}|m{0.47\textwidth}|}
\hline
\centering\textbf{Split by \texttt{\textbackslash n\textbackslash n}} & \centering\textbf{LLM-based self-segmentation (Chunks-of-Thought)} \tabularnewline
\hline

\textbf{chunk 1} \par
Let's start by analyzing the given conditions:
1. When 5655 is divided by \( N \), the remainder is 11. This can be written as:
   \[
   5655 = qN + 11
   \]
   where \( q \) is some integer.
   
\vspace{6pt}

\textbf{chunk 2} \par
2. When 5879 is divided by \( N \), the remainder is 14. This can be written as:
   \[
   5879 = mN + 14
   \]
   where \( m \) is some integer.

\vspace{6pt}
\textbf{chunk 3} \par
We can express these equations as:
\[
5655 - 11 = 5644 = Nq
\]
\[
5879 - 14 = 5865 = Nm
\]

\vspace{6pt}
\textbf{chunk 4} \par
We know that \( N \) is a two-digit positive integer. Therefore, we need to find a value of \( N \) that satisfies the above conditions for integer \( q \) and \( m \).

\vspace{6pt}
\textbf{chunk 5} \par
First, let's determine the possible values of \( N \) by factoring 5644 and 5865:
\[
5644 = 2^2 \cdot 7 \cdot 11^2
\]
\[
5865 = 3 \cdot 5 \cdot 13^2
\]

\vspace{6pt}
\textbf{chunk 6} \par
We need to find a common divisor for both \( 5644 \) and \( 5865 \). Let's check the common divisors:
\[
\text{Common divisors are: } 1, 7, 11, 143
\]

\vspace{6pt}
\textbf{chunk 7} \par
Since \( N \) is a two-digit positive integer, we consider the possible values:
\[
N = 11 \quad \text{(if 11 is a divisor of both 5644 and 5865)}
\]

\vspace{6pt}
\textbf{chunk 8} \par
Let's test \( N = 11 \):
1. Check if 11 divides 5644 exactly:
   \[
   5644 = 11 \times 513 \quad \text{Yes}
   \]
2. Check if 11 divides 5865 exactly:
   \[
   5865 = 11 \times 533 \quad \text{Yes}
   \]

\vspace{6pt}
\textbf{chunk 9} \par
Since \( N = 11 \) divides 5644 and 5865 exactly, \( N = 11 \) satisfies the conditions. Now we have:
\[
Nq = 5644 \quad \text{and} \quad Nm = 5865
\]

\vspace{6pt}
\textbf{chunk 10} \par
Using the Law of Cosines (not needed for this case, but for completeness):
\[
\cos(\theta) = \frac{a^2 + b^2 - c^2}{2ab}
\]
where \( a = 5644 \), \( b = 5865 \), and \( c \) is the hypotenuse (since 11 * 11 * 1 is our \( N \)).

\vspace{6pt}
\textbf{chunk 11} \par
Sum of the digits of \( N \):
\[
1 + 1 = 2
\]

\vspace{6pt}
\textbf{chunk 12} \par
Thus, the sum of the digits of \( N \) is:
\[
\boxed{2}
\]

&

\textbf{chunk 1 (Formulating the Equations)} \par
Let's start by analyzing the given conditions:
1. When 5655 is divided by \( N \), the remainder is 11. This can be written as:
   \[
   5655 = qN + 11
   \]
   where \( q \) is some integer.

2. When 5879 is divided by \( N \), the remainder is 14. This can be written as:
   \[
   5879 = mN + 14
   \]
   where \( m \) is some integer.

We can express these equations as:
\[
5655 - 11 = 5644 = Nq
\]
\[
5879 - 14 = 5865 = Nm
\]

\vspace{6pt}

\textbf{chunk 2 (Finding Common Divisors via Factorization)} \par
We know that \( N \) is a two-digit positive integer. Therefore, we need to find a value of \( N \) that satisfies the above conditions for integer \( q \) and \( m \).

First, let's determine the possible values of \( N \) by factoring 5644 and 5865:
\[
5644 = 2^2 \cdot 7 \cdot 11^2
\]
\[
5865 = 3 \cdot 5 \cdot 13^2
\]

We need to find a common divisor for both \( 5644 \) and \( 5865 \). Let's check the common divisors:
\[
\text{Common divisors are: } 1, 7, 11, 143
\]

\vspace{6pt}

\textbf{chunk 3 (Verifying the Candidate N)} \par
Since \( N \) is a two-digit positive integer, we consider the possible values:
\[
N = 11 \quad \text{(if 11 is a divisor of both 5644 and 5865)}
\]

Let's test \( N = 11 \):
1. Check if 11 divides 5644 exactly:
   \[
   5644 = 11 \times 513 \quad \text{Yes}
   \]
2. Check if 11 divides 5865 exactly:
   \[
   5865 = 11 \times 533 \quad \text{Yes}
   \]

Since \( N = 11 \) divides 5644 and 5865 exactly, \( N = 11 \) satisfies the conditions.

\vspace{6pt}

\textbf{chunk 4 (Final Calculation)} \par
Now we have:
\[
Nq = 5644 \quad \text{and} \quad Nm = 5865
\]

Using the Law of Cosines (not needed for this case, but for completeness):
\[
\cos(\theta) = \frac{a^2 + b^2 - c^2}{2ab}
\]
where \( a = 5644 \), \( b = 5865 \), and \( c \) is the hypotenuse (since 11 * 11 * 1 is our \( N \)).

Sum of the digits of \( N \):
\[
1 + 1 = 2
\]

Thus, the sum of the digits of \( N \) is:
\[
\boxed{2}
\]
\tabularnewline
\hline
\end{tabular}
\end{table*}

\begin{table}[t!]
\small
  \centering
    \caption{An illustrative example from the Process-Bench dataset, where the ground-truth label is ``correct.'' The judge model is \method{} with \texttt{Rel-Ratio} and Qwen2.5-7B-chunk. }
\begin{tcolorbox}[
    colback=gray!5,
    colframe=gray!60!black,
    title=Example,
    fonttitle=\bfseries,
    sharp corners,
    boxrule=1pt,
]
\textbf{Mathematical Problem}\\ Cindy's math and science books weigh 2 pounds each. Her French book weighs 4 pounds and her English book weighs 3 pounds. Her history book weighs twice as much as her English book. If Cindy carries all of her books at once, what will be the total weight of the books she is carrying? \\

\vspace{0.5em}\hrule\vspace{1em}
\textbf{Historical Reasoning Path}
\begin{itemize}
    \item To determine the total weight of all Cindy's books, we need to calculate the weight of each book individually and then sum these weights.
    \item \textbf{Math and Science Books:} \\
    - Each math book weighs 2 pounds. \\
    - Each science book weighs 2 pounds. \\
    - Cindy has 2 math books and 2 science books. \\
    - Total weight of math books: \(2 \times 2 = 4\) pounds. \\
    - Total weight of science books: \(2 \times 2 = 4\) pounds. \\
    - Combined weight: \(4 + 4 = 8\) pounds.
    \item \textbf{French Book:} \\
    - The French book weighs 4 pounds.
\end{itemize}

\vspace{0.5em}\hrule\vspace{1em}
\textbf{New Reasoning Chunk} \\
Third, for the English book: 
\begin{itemize}
    \item The English book weighs 3 pounds. 
\end{itemize}
\vspace{0.5em}\hrule\vspace{1em}
\textbf{Judgment}\\ To determine if the "New Reasoning\_chunk" contains any explicit errors, we need to analyze the information given in the problem and the reasoning presented. \\
The new reasoning chunk states: 
Third, for the English book: 
\begin{itemize}
    \item The English book weighs 3 pounds. 
\end{itemize}
From the problem statement, we know: 
\begin{itemize}
    \item The French book weighs 4 pounds.
    \item The English book weighs 3 pounds. 
    \item The history book weighs twice as much as the English book, so the history book weighs \(2 \times 3 = 6\) pounds. 
    \item The math and science books each weigh 2 pounds, and Cindy has 2 math books and 2 science books.
\end{itemize}
The new reasoning chunk simply states that the English book weighs 3 pounds, which is directly given in the problem statement. Therefore, this statement is correct and logical. Thus, the final judgment is: \[ \boxed{Positive} \] .

\label{tab:judge_example_corr1}\vspace{2pt}
\end{tcolorbox}
\end{table}

\begin{table}[t!]
\small
  \centering
    \caption{An illustrative example from the Process-Bench dataset, where the ground-truth label is ``correct.'' The judge model is \method{} with \texttt{Rel-Ratio} and Qwen2.5-7B-chunk. }
\begin{tcolorbox}[
    colback=gray!5,
    colframe=gray!60!black,
    title=Example,
    fonttitle=\bfseries,
    sharp corners,
    boxrule=1pt,
]
\small
\textbf{Mathematical Problem}\\ 
An $\boldsymbol{n}$-label is a permutation of the numbers 1 through $n$. For example, $J=35214$ is a 5-label and $K=132$ is a 3-label. For a fixed positive integer $p$, where $p \leq n$, consider consecutive blocks of $p$ numbers in an $n$-label. For example, when $p=3$ and $L=263415$, the blocks are 263, 634, 341, and 415. We can associate to each of these blocks a $p$-label that corresponds to the relative order of the numbers in that block. For $L=263415$, we get the following:
\[
\underline{263}\,415 \rightarrow 132 ; \quad 2\underline{634}15 \rightarrow 312 ; \quad 26\underline{341}5 \rightarrow 231 ; \quad 263\underline{415} \rightarrow 213
\]
Moving from left to right in the $n$-label, there are $n-p+1$ such blocks, which means we obtain an $(n-p+1)$-tuple of $p$-labels. For $L=263415$, we get the 4-tuple $(132, 312, 231, 213)$. We will call this $(n-p+1)$-tuple the $\boldsymbol{p}$-signature of $L$ (or signature, if $p$ is clear from the context) and denote it by $S_p[L]$; the $p$-labels in the signature are called windows. For $L=263415$, the windows are 132, 312, 231, and 213, and we write
\[
S_3[263415] = (132, 312, 231, 213)
\]
More generally, we will call any $(n-p+1)$-tuple of $p$-labels a $p$-signature, even if we do not know of an $n$-label to which it corresponds (and even if no such label exists). A signature that occurs for exactly one $n$-label is called \emph{unique}, and a signature that doesn't occur for any $n$-labels is called \emph{impossible}. A \emph{possible} signature is one that occurs for at least one $n$-label.

In this power question, you will be asked to analyze some of the properties of labels and signatures.

\textbf{Compute the 3-signature for 52341.}
\vspace{0.5em}\hrule\vspace{1em}
\textbf{Historical Reasoning Path}\\
To compute the 3-signature of the 5-label \( L = 52341 \), we need to identify all the consecutive blocks of 3 numbers in the label and then convert each block into its corresponding 3-label based on the relative order of the numbers.\\

First, identify the first block: \( 523 \). The relative order of the numbers 5, 2, and 3 is 3, 1, and 2 respectively. Therefore, the 3-label for this block is \( 312 \).\\

Second, identify the second block: \( 234 \). The relative order of the numbers 2, 3, and 4 is 1, 2, and 3 respectively. Therefore, the 3-label for this block is \( 123 \).

\vspace{0.5em}\hrule\vspace{1em}
\textbf{New Reasoning Chunk} \\
Third, identify the third block: \( 341 \). The relative order of the numbers 3, 4, and 1 is 2, 3, and 1 respectively. Therefore, the 3-label for this block is \( 231 \).
\vspace{0.5em}\hrule\vspace{1em}
\textbf{Judgment}\\ 
To determine if the "New Reasoning\_chunk" contains any explicit errors, we need to verify the logic for identifying the 3-label of the block \(341\) in the 5-label \(52341\).

The new reasoning chunk states:

\begin{quote}
Third, identify the third block: \(341\). The relative order of the numbers 3, 4, and 1 is 2, 3, and 1 respectively. Therefore, the 3-label for this block is \(231\).
\end{quote}

Let's check the relative order of the numbers in the block \(341\):
\begin{itemize}
  \item The number 3 is greater than 1 but less than 4.
  \item The number 4 is greater than both 3 and 1.
  \item The number 1 is the smallest.
\end{itemize}

Thus, the relative order of the numbers 3, 4, and 1 is indeed 2, 3, and 1 respectively. This means the 3-label for the block \(341\) is \(231\).\\

Since the reasoning follows correctly from the definition of a 3-label, the "New Reasoning\_chunk" does not contain any explicit errors.\\

Therefore, the final judgment is:

\[
\boxed{\text{Positive}}
\]

\label{tab:judge_example_corr2}\vspace{2pt}
\end{tcolorbox}
\end{table}

\begin{table}[t!]
\small
  \centering
    \caption{An illustrative example from the Process-Bench dataset, where the ground-truth label is ``incorrect.'' The judge model is \method{} with \texttt{Rel-Ratio} and Qwen2.5-7B-chunk. We highlight the key parts of the reasoning trajectory in red. }
\begin{tcolorbox}[
    colback=gray!5,
    colframe=gray!60!black,
    title=Example,
    fonttitle=\bfseries,
    sharp corners,
    boxrule=1pt,
]
\textbf{Mathematical Problem}\\ 
There are 60 empty boxes $B_1,\ldots,B_{60}$ in a row on a table and an unlimited supply of pebbles. Given a positive integer $n$, Alice and Bob play the following game.\\
In the first round, Alice takes $n$ pebbles and distributes them into the 60 boxes as she wishes. Each subsequent round consists of two steps:\\
(a) Bob chooses an integer $k$ with $1\leq k\leq 59$ and splits the boxes into the two groups $B_1,\ldots,B_k$ and $B_{k+1},\ldots,B_{60}$.\\
(b) Alice picks one of these two groups, adds one pebble to each box in that group, and removes one pebble from each box in the other group.\\
Bob wins if, at the end of any round, some box contains no pebbles. Find the smallest $n$ such that Alice can prevent Bob from winning.

\vspace{0.5em}\hrule\vspace{1em}
\textbf{Historical Reasoning Path}\\
To find the smallest \( n \) such that Alice can prevent Bob from winning, we need to understand the constraints imposed by Bob's moves and how Alice can counter them. Let's consider the minimum number of pebbles each box must have after the first round. Since there are 60 boxes and Alice distributes \( n \) pebbles among them, the minimum number of pebbles each box will get is \( \left\lfloor \frac{n}{60} \right\rfloor \), where \( \left\lfloor x \right\rfloor \) represents the greatest integer less than or equal to \( x \).

\vspace{0.5em}\hrule\vspace{1em}
\textbf{New Reasoning Chunk} \\
Now, let's analyze Bob's move in each round. In step (a), he splits the boxes into two groups of size \( k \) and \( 60 - k \). In step (b), Alice adds a pebble to each box in the chosen group and removes a pebble from each box in the other group. For Alice to prevent Bob from winning, each box must have at least one pebble after her move. \textcolor{red}{This means that the difference between the total number of pebbles in the chosen group and the other group should be at most 1.} \vspace{0.5em}\hrule\vspace{1em}
\textbf{Judgment}\\ 
Let's break this down step by step:\\
\begin{enumerate}
    \item **Understanding Bob's Move**: In step (a), Bob splits the boxes into two groups of size \( k \) and \( 60 - k \). In step (b), Alice adds one pebble to each box in the chosen group and removes one pebble from each box in the other group. This means that the total number of pebbles in the chosen group increases by \( k \) and the total number of pebbles in the other group decreases by \( 60 - k \).
    \item **Condition for Alice to Prevent Bob from Winning**: For Alice to prevent Bob from winning, each box must have at least one pebble after her move. However, the statement "the difference between the total number of pebbles in the chosen group and the other group should be at most 1" is not necessarily true. The correct condition would be that the difference in the total number of pebbles between the two groups after Alice's move should not cause any box to have zero pebbles.
\end{enumerate}
\textcolor{red}{The reasoning in the new reasoning chunk assumes that the difference in the number of pebbles between the two groups should be at most 1, which is not necessarily the case.} The correct condition would be that after Alice's move, the number of pebbles in each box should be at least 1, not necessarily that the difference in the total number of pebbles between the two groups should be at most 1.\\

Therefore, the "New Reasoning\_chunk" contains explicit errors. The correct interpretation of the condition for Alice to prevent Bob from winning is not correctly stated in the new reasoning chunk.\\

The final judgment is:
\[
\boxed{\text{Negative}}
\]

\label{tab:judge_example_err1}\vspace{2pt}
\end{tcolorbox}
\end{table}

\begin{table}[t!]
\small
  \centering
    \caption{An illustrative example from the Process-Bench dataset, where the ground-truth label is ``incorrect.'' The judge model is \method{} with \texttt{Rel-Ratio} and Qwen2.5-7B-chunk. We highlight the key parts of the reasoning trajectory in red.}
\begin{tcolorbox}[
    colback=gray!5,
    colframe=gray!60!black,
    title=Example,
    fonttitle=\bfseries,
    sharp corners,
    boxrule=1pt,
]
\textbf{Mathematical Problem}\\ 
Let $n$ and $k$ be fixed positive integers of the same parity, $k \geq n$. We are given $2 n$ lamps numbered 1 through $2 n$; each of them can be on or off. At the beginning all lamps are off. We consider sequences of $k$ steps. At each step one of the lamps is switched (from off to on or from on to off).\\

Let $N$ be the number of $k$-step sequences ending in the state: lamps $1, \ldots, n$ on, lamps $n+1, \ldots, 2 n$ off.\\

Let $M$ be the number of $k$-step sequences leading to the same state and not touching lamps $n+1, \ldots, 2 n$ at all.\\

Find the ratio $N / M$.

\vspace{0.5em}\hrule\vspace{1em}
\textbf{Historical Reasoning Path}\\
To solve the problem, we need to determine the number of \( k \)-step sequences that end in the state where lamps \( 1, \ldots, n \) are on and lamps \( n+1, \ldots, 2n \) are off, and compare it to the number of such sequences that do not touch lamps \( n+1, \ldots, 2n \) at all.

\vspace{0.5em}\hrule\vspace{1em}
\textbf{New Reasoning Chunk} \\
First, we determine \( N \). The total number of lamps is \( 2n \), and we need exactly \( n \) of them to be on and \( n \) to be off. Since the sequence ends with lamps \( 1, \ldots, n \) on and \( n+1, \ldots, 2n \) off, we need to switch exactly \( n \) times among the first \( n \) lamps and the remaining \( k - n \) times among the last \( n \) lamps. However, since the last \( n \) lamps are always off, these \( k - n \) switches do not affect their state. Therefore, we only need to count the number of ways to choose \( n \) steps out of \( k \) to switch the first \( n \) lamps. \textcolor{red}{This is given by the binomial
coefficient:}
\[ N = \binom{k}{n}. \]
\vspace{0.5em}\hrule\vspace{1em}
\textbf{Judgment}\\ 
To determine if the "New Reasoning\_chunk" contains any explicit errors, we need to carefully analyze the statement and check its logic step by step.\\

The reasoning chunk suggests that \( N \), the number of \( k \)-step sequences ending in the state where lamps \( 1, \ldots, n \) are on and lamps \( n+1, \ldots, 2n \) are off, \textcolor{red}{is given by the binomial coefficient \( \binom{k}{n} \). However, this conclusion is not necessarily correct because it does not take into account the specific sequence of switching the lamps.}\\

The correct approach to find \( N \) is to consider that we need exactly \( n \) switches among the first \( n \) lamps to turn them on, and the remaining \( k-n \) switches can be any combination of the \( n \) lamps (including possibly switching some lamps on and off multiple times, but ensuring that at the end, each of the first \( n \) lamps is on). The number of such sequences is not simply \( \binom{k}{n} \), because it does not account for the specific order in which the switches occur.\\

In fact, the correct way to calculate \( N \) is more complex and involves considering the parity of \( k \) and \( n \) (which are the same) and the specific sequences of switching. The reasoning chunk simplifies the problem incorrectly by assuming that any \( n \) switches among the first \( n \) lamps will result in all of them being on, without considering the constraints of the problem.\\

Therefore, the "New Reasoning\_chunk" contains explicit errors. The correct reasoning would involve a more detailed analysis of the sequences of switches, taking into account the parity and the specific requirements of the problem.\\

The final judgment is:
\[
\boxed{\text{Negative}}
\]
\label{tab:judge_example_err2}\vspace{2pt}
\end{tcolorbox}
\end{table}

\begin{table}[t!]
\small
  \centering
    \caption{An illustrative example from the Process-Bench dataset, where the ground-truth label is ``incorrect.'' The judge model is \method{} with \texttt{Rel-Ratio} and Qwen2.5-7B-chunk. We highlight the key parts of the reasoning trajectory in red.}
\begin{tcolorbox}[
    colback=gray!5,
    colframe=gray!60!black,
    title=Example,
    fonttitle=\bfseries,
    sharp corners,
    boxrule=1pt,
]
\textbf{Mathematical Problem}\\ 
Determine all functions $f$ from the set of positive integers into the set of positive integers such that for all $x$ and $y$ there exists a non degenerated triangle with sides of lengths

$$
x, \quad f(y) \text { and } f(y+f(x)-1) .
$$
\vspace{0.5em}\hrule\vspace{1em}
\textbf{Historical Reasoning Path}\\
To determine all functions \( f \) from the set of positive integers into the set of positive integers such that for all \( x \) and \( y \) there exists a non-degenerate triangle with sides of lengths \( x \), \( f(y) \), and \( f(y+f(x)-1) \), we need to use the triangle inequality. The triangle inequality states that the sum of the lengths of any two sides of a triangle must be greater than the length of the remaining side. Therefore, we have the following three inequalities: First, \( x + f(y) > f(y+f(x)-1) \)\
Second, \( x + f(y+f(x)-1) > f(y) \)\\
Third, \( f(y) + f(y+f(x)-1) > x \)\\
We will analyze these inequalities one by one.\\
\#\#\# Step 1: Analyze the inequality \( x + f(y) > f(y+f(x)-1) \)\\
This inequality can be rewritten as:\\
\[ f(y+f(x)-1) < x + f(y) \]
This tells us that \( f(y+f(x)-1) \) is strictly less than \( x + f(y) \).

\vspace{0.5em}\hrule\vspace{1em}
\textbf{New Reasoning Chunk} \\
\#\#\# Step 2: Analyze the inequality \( x + f(y+f(x)-1) > f(y) \)\\
This inequality can be rewritten as:
\[ f(y+f(x)-1) > f(y) - x \]
This tells us that \( f(y+f(x)-1) \) is strictly greater than \( f(y) - x \). Since \( f(y) \) and \( x \) are positive integers, \( f(y) - x \) could be negative or zero, but \( f(y+f(x)-1) \) must be a positive integer. Therefore, the smallest possible value for \( f(y) - x \) is 0, \textcolor{red}{which means \( f(y) \geq x \).}
\vspace{0.5em}\hrule\vspace{1em}
\textbf{Judgment}\\ 
To determine if the "New Reasoning\_chunk" contains any explicit errors, we need to analyze the given inequality \( x + f(y+f(x)-1) > f(y) \) and see if it leads to the conclusion that \( f(y) \geq x \).\\

Starting with the inequality:
\[ x + f(y+f(x)-1) > f(y) \]
we can rearrange it to:
\[ f(y+f(x)-1) > f(y) - x \]
Since \( f(y) \) and \( x \) are positive integers, \( f(y) - x \) could be negative or zero. However, \( f(y+f(x)-1) \) must be a positive integer. This means that \( f(y) - x \) must be less than \( f(y+f(x)-1) \), but it does not necessarily mean that \( f(y) \geq x \). \textcolor{red}{The inequality \( f(y+f(x)-1) > f(y) - x \) only tells us that \( f(y) - x \) is less than a positive integer, which does not imply that \( f(y) \) itself must be at least \( x \).}\\

Therefore, the reasoning in the new reasoning chunk contains an explicit error in its conclusion that \( f(y) \geq x \). The correct interpretation of the inequality \( f(y+f(x)-1) > f(y) - x \) does not necessarily lead to \( f(y) \geq x \).\\

The final judgment is:
\[
\boxed{\text{Negative}}
\]
\label{tab:judge_example_err3}\vspace{2pt}
\end{tcolorbox}
\end{table}

\begin{table}[t!]
\scriptsize
  \centering
    \caption{An illustrative example from the Process-Bench dataset, where the ground-truth label is ``incorrect.'' The judge model is \method{} with \texttt{Rel-Ratio} and Qwen2.5-7B-chunk. We highlight the key parts of the reasoning trajectory in red.}
\begin{tcolorbox}[
    colback=gray!5,
    colframe=gray!60!black,
    title=Example,
    fonttitle=\bfseries,
    sharp corners,
    boxrule=1pt,
]
\textbf{Mathematical Problem}\\ 
How many ways are there to put 4 balls in 3 boxes if the balls are distinguishable but the boxes are not?
\vspace{0.5em}\hrule\vspace{1em}
\textbf{Historical Reasoning Path}\\
To solve this problem, we need to consider the number of ways to distribute 4 distinguishable balls into 3 indistinguishable boxes. \\

Step 1: Counting the total number of distributions if the boxes were distinguishable. If the boxes were distinguishable, each ball would have 3 choices (3 boxes). Since there are 4 distinguishable balls, the total number of distributions would be $3^4 = 81$. However, since the boxes are indistinguishable, we need to adjust for overcounting.\\

Step 2: Identifying the cases where the boxes are indistinguishable. There are three main cases where the boxes are indistinguishable:\\

First, all 4 balls in one box (1 way).\\
Second, 3 balls in one box and 1 ball in another box (4 ways, since the balls are distinguishable and we can choose any of the 4 balls to be alone).\\
Third, 2 balls in one box and 2 balls in another box (6 ways, since there are 6 pairs of balls that can be in the same box). However, these cases also include some overcounting since the boxes are indistinguishable. We need to adjust for this overcounting.

\vspace{0.5em}\hrule\vspace{1em}
\textbf{New Reasoning Chunk} \\
Step 3: Adjusting for overcounting. In case 2, when we have 3 balls in one box and 1 ball in another box, we counted each distribution 3 times (since there are 3 boxes and we could swap the lone ball with the other balls). So, we divide by 3 for this case. In case 3, when we have 2 balls in one box and 2 balls in another box, we counted each distribution 6/2 = 3 times (since there are 3 pairs of boxes). So, we divide by 3 for this case. Now we have:
\begin{itemize}
    \item  Case 1: 1 way
    \item \textcolor{red}{Case 2: 4 / 3 ways}
    \item Case 3: 6 / 3 ways
\end{itemize}

\vspace{0.5em}\hrule\vspace{1em}
\textbf{Judgment}\\ 
To determine if the "New Reasoning\_chunk" contains any explicit errors, we need to carefully analyze the reasoning step by step.

The new reasoning chunk states: In case 2, when we have 3 balls in one box and 1 ball in another box, we counted each distribution 3 times (since there are 3 boxes and we could swap the lone ball with the other balls). So, we divide by 3 for this case. In case 3, when we have 2 balls in one box and 2 balls in another box, we counted each distribution 6/2 = 3 times (since there are 3 pairs of boxes). So, we divide by 3 for this case...\\

Let's evaluate each case:\\

\textbf{Case 2: 3 balls in one box and 1 ball in another box}\\
When we have 3 balls in one box and 1 ball in another box, we do indeed count each distribution 3 times because we can place the single ball in any one of the 3 boxes. Therefore, the correct number of distinct distributions for this case is \(4\) (the number of ways to choose which ball will be the single ball) divided by \(3\), \textcolor{red}{but this division is incorrect because it assumes that each distribution is counted exactly 3 times}, which is not necessarily true. The correct approach would be to consider the different ways to assign the balls without overcounting due to the indistinguishability of the boxes. The correct count for this case is actually 4, not \(4/3\).

\textbf{Case 3: 2 balls in one box and 2 balls in another box}\\
When we have 2 balls in one box and 2 balls in another box, we need to count the number of ways to partition the 4 balls into two groups of 2. The number of ways to choose 2 balls out of 4 is \(\binom{4}{2} = 6\), but since the boxes are indistinguishable, we have counted each distribution twice (once for each way to assign the groups to the boxes). Therefore, the correct number of distinct distributions for this case is \(6 / 2 = 3\), which the reasoning chunk correctly states.\\

Given the above analysis, the reasoning in the new reasoning chunk contains an explicit error in its calculation for Case 2. Therefore, the final judgment is:

\[
\boxed{\text{Negative}}
\]
\label{tab:judge_example_err4}\vspace{2pt}
\end{tcolorbox}
\end{table}

\newpage

\begin{table}[t!]
\scriptsize
  \centering
    \caption{An illustrative example of chunk-reset reasoning. The judge model is \method{} with \texttt{Rel-Effective} and Qwen2.5-7B-chunk. In the first attempt of chunk 3, it contains a significant algebraic mistake in the formula relating $\lambda$ and $r$. }
\begin{tcolorbox}[
    colback=gray!5,
    colframe=gray!60!black,
    title=Example,
    fonttitle=\bfseries,
    sharp corners,
    boxrule=1pt,
]
\textbf{Mathematical Problem}\\ 
8 singers participate in the arts festival, preparing to arrange $m$ performances, each time 4 of them will perform on stage, requiring that any two of the 8 singers perform together the same number of times. Please design a scheme so that the number of performances $m$ is the least.
\vspace{0.5em}\hrule\vspace{1em}
\textbf{Chunk 1}\\
To solve this problem, we need to ensure that each pair of singers performs together in the same number of performances. This is a specific type of combinatorial design problem, and the solution can be approached using the concept of a balanced incomplete block design (BIBD). \\ \\ Here are the steps to find the solution:
\textbf{1. Identify the parameters of the BIBD:}
\begin{itemize}
    \item Let ( v = 8 ) (the number of singers).
    \item Let ( k = 4 ) (the number of singers performing in each performance).
    \item Let ( r ) be the number of performances each singer participates in.
    \item Let ( $\lambda$ ) be the number of performances each pair of singers participates in together.
    \item Let ( b ) be the number of performances (the number we need to determine).
\end{itemize}

\vspace{0.5em}\hrule\vspace{1em}
\textbf{Chunk 2}\\
\textbf{2. Use the BIBD parameters to derive equations:}  The equation for the number of performances each singer participates in is ( r(k-1) = (v-1)$\lambda$ ). The equation for the number of performances is $( b = \frac{vr}{k} )$.

\vspace{0.5em}\hrule\vspace{1em}
\textbf{Chunk 3} (first attempt):  \textcolor{red}{This step uses the incorrect formula $\lambda = \frac{4(r-1)}{7}$ instead of the correct one $\lambda = \frac{3r}{7}$.} \\
\textbf{3. Determine the value of ($\lambda$):} 
\begin{itemize}
    \item From the first equation, we can express ($\lambda$) as ($\lambda = \frac{r(k-1)}{v-1}$).
    \item Substituting ( v = 8 ) and ( k = 4 ), we get ($\lambda = \frac{4r-4}{7} = \frac{4(r-1)}{7}$).
    \item Since ($\lambda$) must be an integer, ( 4(r-1) ) must be divisible by 7. The smallest positive integer ( r ) that satisfies this condition is ( r = 2 ) (since ( 4(2-1) = 4 ), and 4 is not divisible by 7, the next is ( r = 8 ) which gives ( 4(8-1) = 28 ), which is divisible by 7).
    \item  Therefore, ($\lambda$ = $\frac{28}{7} = 4$).
\end{itemize}

\vspace{0.5em}\hrule\vspace{1em}
\textbf{Chunk 3} (second attempt)
\\
\textbf{3.Calculate the parameters:}
\begin{itemize}
    \item First, we need to find ($\lambda$). We know that $( r = \frac{(v-1)\lambda}{k-1} )$. For $(\lambda)$ to be an integer, $(v-1)\lambda$ must be divisible by $(k-1 = 3)$. So, $\lambda$ must be a multiple of 3.
    \item Let's try ($\lambda$ = 1). Then $( r = \frac{(8-1) \cdot 1}{4-1} = \frac{7}{3} )$, which is not an integer.
    \item Let's try ($\lambda$ = 3). Then $( r = \frac{(8-1) \cdot 3}{4-1} = \frac{21}{3} = 7 )$.
    \item Now, we can calculate ( b ): $( b = \frac{8 \cdot 7}{4} = 14 )$.
\end{itemize}

\vspace{0.5em}\hrule\vspace{1em}
\textbf{Chunk 4}\\
\textbf{4. Verify the design:}
\begin{itemize}
    \item We need to check if we can indeed arrange 14 performances with 8 singers, 4 at a time, such that each pair of singers performs together exactly 3 times.
    \item his can be verified using known designs or by constructing a specific design, but for the sake of this problem, we assume the existence of such a design.
\end{itemize}
Therefore, the least number of performances m  is $$\boxed{14}.$$

\label{tab:chunk_reset_example_1}\vspace{2pt}
\end{tcolorbox}
\end{table}

\newpage
\begin{table}[t!]
\tiny
  \centering
    \caption{An illustrative example of chunk-reset reasoning. The judge model is \method{} with \texttt{Rel-Effective} and Qwen2.5-7B-chunk. In the first attempt of chunk 3, the model fails to verify the condition. }
\begin{tcolorbox}[
    colback=gray!5,
    colframe=gray!60!black,
    title=Example,
    fonttitle=\bfseries,
    sharp corners,
    boxrule=1pt,
]
\textbf{Mathematical Problem}\\ 
Which is the two-digit number that, when divided by the digit in the units place, gives a quotient of 9 and a remainder of 6?
\vspace{0.5em}\hrule\vspace{1em}
\textbf{Chunk 1}\\
To solve the problem, let's break it down step by step.

\textbf{1. Identify the structure of the number and the division operation:}
\begin{itemize}
    \item Let the two-digit number be represented as 10a + b, where a is the tens digit and b is the units digit.
    \item According to the problem, when this number is divided by the digit in the units place (which is b), the quotient is 9 and the remainder is 6. This can be expressed as:
    $$ 10a + b = 9b + 6 $$
\end{itemize}

\vspace{0.5em}\hrule\vspace{1em}
\textbf{Chunk 2}\\
\textbf{2. **Simplify the equation:}
\begin{itemize}
    \item Rearrange the equation to isolate terms involving a:
    $$10a + b - 9b = 6$$
    \item Simplify:$ 10a - 8b = 6$
    \item Further simplify by dividing the entire equation by 2: $5a - 4b = 3$
\end{itemize}

\vspace{0.5em}\hrule\vspace{1em}
\textbf{Chunk 3} (first attempt) \textcolor{red}{This step fails verify the condition and stop with $b=3$.}\\
 \textbf{3.Solve for a and b:}
 \begin{itemize}
     \item We need to find integer values of a and b that satisfy the equation 5a - 4b = 3. Since a and b are digits (0-9) and a cannot be 0 (as 10a would not be a two-digit number), let’s test possible values for b.
     \item If b = 2: 
     $$ 5a - 4(2) = 3 \implies 5a - 8 = 3 \implies 5a = 11$$ 
     5a = 11 does not give an integer value for a.
     \item If b = 1:  
     $$ 5a - 4(1) = 3 \implies 5a - 4 = 3 \implies 5a = 7$$ 
     5a = 7 does not give an integer value for a.
     \item If b = 3:
     $$ 5a - 4(3) = 3 \implies 5a - 12 = 3 \implies 5a = 15 \implies a = 3$$ 
     This gives an integer value for a.
 \end{itemize}

\vspace{0.5em}\hrule\vspace{1em}
\textbf{Chunk 3} (second attempt)\\
\textbf{**Determine possible values for a and b:}
\begin{itemize}
    \item Since a and b are digits (0-9), we need to find values that satisfy the equation 5a - 4b = 3.
    \item We will test possible values for b (since b is the units digit and must be a digit from 1 to 9, as 0 would make the remainder 6 impossible). 
    \begin{itemize}
        \item Testing b = 1:
    $$5a - 4(1) = 3 \implies 5a - 4 = 3 \implies 5a = 7 \implies a = \frac{7}{5} \quad (\text{not an integer})$$
    \item Testing b = 3: $$5a - 4(3) = 3 \implies 5a - 12 = 3 \implies 5a = 15 \implies a = 3$$
    a = 3 is an integer, and we need to check if b = 3 is feasible. If a = 3 and b = 3, the number is 33. We need to verify:
    $$33 \div 3 = 11 \quad (\text{quotient} = 11, \text{remainder} = 0) \quad (\text{not a solution})$$
    \item Testing b = 4:
    $5a - 4(4) = 3 \implies 5a - 16 = 3 \implies 5a = 19 \implies a = \frac{19}{5} \quad (\text{not an integer})$
    \item Testing b = 5:$ 5a - 4(5) = 3 \implies 5a - 20 = 3 \implies 5a = 23 \implies a = \frac{23}{5} \quad (\text{not an integer})$
    \item Testing b = 6: $ 5a - 4(6) = 3 \implies 5a - 24 = 3 \implies 5a = 27 \implies a = \frac{27}{5} \quad (\text{not an integer})$
    \item Testing b = 7: $ 5a - 4(7) = 3 \implies 5a - 28 = 3 \implies 5a = 31 \implies a = \frac{31}{5} \quad (\text{not an integer})$
    \item Testing b = 8: $ 5a - 4(8) = 3 \implies 5a - 32 = 3 \implies 5a = 35 \implies a = 7.$
    a = 7 is an integer. So, the number is 78. We need to verify:
    $$ 78 \div 8 = 9 \quad (\text{quotient} = 9, \text{remainder} = 6)
    $$ This satisfies the condition.
    \end{itemize}
    
\end{itemize}

\vspace{0.5em}\hrule\vspace{1em}
\textbf{Chunk 4}\\
\textbf{4. Conclusion:}
The two-digit number that meets the condition is 78. Therefore, the final answer is:
$$\boxed{78}$$

\label{tab:chunk_reset_example_2}\vspace{2pt}
\end{tcolorbox}
\end{table}

\newpage
\begin{table}[t!]
\tiny
  \centering
    \caption{An illustrative example of chunk-reset reasoning. The judge model is \method{} with \texttt{Rel-Effective} and Qwen2.5-7B-chunk. The idea is to use the commutative law of multiplication and division to perform large-scale reduction, while the rejected chunks focus on the approximation for simplifying the expression.}
\begin{tcolorbox}[
    colback=gray!5,
    colframe=gray!60!black,
    title=Example,
    fonttitle=\bfseries,
    sharp corners,
    boxrule=1pt,
]
\textbf{Mathematical Problem}\\ 
Calculate: $241 \times 345 \div 678 \div 345 \times(678 \div 241)$
\vspace{0.5em}\hrule\vspace{1em}
\textbf{Chunk 1 (first attempt)}  \textcolor{red}{we don't need the approximation for simplification}\\
Let's break down the given expression step by step: 
$$241 \times 345 \div 678 \div 345 \times \left( \frac{678}{241} \right).$$
First, let's simplify the expression inside the parentheses:
$$ \left( \frac{678}{241} \right) = 2.812 \approx 2 \text{ (for simplicity, we can approximate it to 2 if dealing with integers)}$$
However, we'll use the exact value to be precise:
$$\left( \frac{678}{241} \right) = 2.812 $$

\vspace{0.5em}\hrule\vspace{1em}
\textbf{Chunk 1 (second attempt)}\\
Let's break down the given expression step by step: 
$$241 \times 345 \div 678 \div 345 \times \left( \frac{678}{241} \right).$$
First, let's simplify the expression by following the operations order:\\
1. Calculate the division inside the parentheses first:
$$ \frac{678}{241} = 2.8 \quad \text{(approximately, but we don't need the approximation for simplification here, just the fraction)}
$$
So the expression becomes:
$$241 \times 345 \div 678 \div 345 \times \frac{678}{241}$$

\vspace{0.5em}\hrule\vspace{1em}
\textbf{Chunk 2} (first attempt) \\
2. Simplify the expression by canceling out common factors:  
\begin{itemize}
    \item The 345 in the numerator and one of the denominators cancel out:
    $$    241 \times 345 \div 678 \div 345 \times \frac{678}{241} = 241 \times \frac{678}{678} \times \frac{678}{241}$$
    \item The 678 in the numerator and one of the denominators cancel out:
    $$    241 \times \frac{678}{678} \times \frac{678}{241} = 241 \times 1 \times \frac{678}{241}$$
    \item The 241 in the numerator and the denominator cancel out:
    $$241 \times 1 \times \frac{678}{241} = 678 \times 1 = 678$$
\end{itemize}

\vspace{0.5em}\hrule\vspace{1em}
\textbf{Chunk 2} (second attempt)\\
2. Notice that $345 \div 345 = 1$, so we can simplify the expression to:
$$241 \times 1 \div 678 \times \frac{678}{241}$$  
$$=241 \div 678 \times \frac{678}{241}$$

\vspace{0.5em}\hrule\vspace{1em}
\textbf{Chunk 3}\\
3. Now, we can cancel out 678 in the numerator and the denominator:
$$ 241 \times \left( \frac{1}{678} \times \frac{678}{241} \right)$$
$$=241 \times \left( \frac{1}{241} \right)$$ 
$$=241 \times \frac{1}{241} = 1$$

\vspace{0.5em}\hrule\vspace{1em}
\textbf{Chunk 4}\\
\label{tab:chunk_reset_example_3}\vspace{2pt}
Therefore, the final answer is:
$$\boxed{1}$$
\end{tcolorbox}
\end{table}

\end{document}